\documentclass[journal]{IEEEtran}
\usepackage{amsmath}
\usepackage{mathtools}
\usepackage{fixmath}
\usepackage{float}
\usepackage{amssymb}
\usepackage{lipsum}
\usepackage{array,multirow,graphicx}
\usepackage[table]{xcolor}
\usepackage{subcaption}
\usepackage{lineno,hyperref}
\usepackage[ruled,longend]{algorithm2e}
\usepackage{booktabs}
\usepackage{hyperref}
\usepackage{rotating}
\hypersetup{
	colorlinks,
	linkcolor=[rgb]{1.0, 0.0, 0.0},
	citecolor=[rgb]{1.0, 0.2, 0.2},
	urlcolor =[rgb]{0.0, 0.0, 0.8}
}
\modulolinenumbers[5]

\usepackage{pifont}% http://ctan.org/pkg/pifont
\ifCLASSINFOpdf
\else
\fi
\begin{document}
	\bstctlcite{IEEEexample:BSTcontrol}
	% to reduce the margin of equations and figure
	%\abovedisplayskip=0pt
	%\abovedisplayshortskip=0pt
	%\belowdisplayskip=0pt
	%%\belowdisplayshortskip=0pt
	%\abovecaptionskip=0pt
	%\belowcaptionskip=0pt
	
	\title{HoVer-Net: Simultaneous Segmentation and Classification of Nuclei in Multi-Tissue Histology Images}
	

	\author{Simon~Graham$^*$, ~Quoc~Dang~Vu$^*$, \\ ~Shan~E~Ahmed~Raza, ~Ayesha Azam, ~Yee Wah Tsang, \\ ~Jin~Tae~Kwak$^+$
		and~Nasir~Rajpoot$^+$% <-this % stops a space
		\thanks{$^*$ First authors contributed equally.}
		\thanks{$^+$ Last authors contributed equally.}
		\thanks{S.Graham, N.Rajpoot and S.E.A.Raza are with the Department of Computer Science, University of Warwick, UK.}
		\thanks{S.Graham is also with the Mathematics for Real-World Systems Centre for Doctoral Training, University of Warwick, UK.}
		\thanks{S.E.A.Raza is also with the Centre for Evolution and Cancer \& Division of Molecular Pathology, The Institute of Cancer Research, London, UK.}
		\thanks{Q.D.Vu and J.T.Kwak are with the Department of Computer Science and Engineering, Sejong University, South Korea.}
		\thanks{A.Azam and Y-W.T are with the Department of Pathology at University Hospitals Coventry and Warwickshire, Coventry, UK}
		}

	\maketitle

\begin{abstract}

    		Nuclear segmentation and classification within Haematoxylin \& Eosin stained histology images is a fundamental prerequisite in the digital pathology work-flow. The development of automated methods for nuclear segmentation and classification enables the quantitative analysis of tens of thousands of nuclei within a whole-slide pathology image, opening up possibilities of further analysis of large-scale nuclear morphometry. However, automated nuclear segmentation and classification is faced with a major challenge in that there are several different types of nuclei, some of them exhibiting large intra-class variability such as the nuclei of tumour cells. Additionally, some of the nuclei are often clustered together. To address these challenges, we present a novel convolutional neural network for simultaneous nuclear segmentation and classification that leverages the instance-rich information encoded within the vertical and horizontal distances of nuclear pixels to their centres of mass. These distances are then utilised to separate clustered nuclei, resulting in an accurate segmentation, particularly in areas with overlapping instances. Then, for each segmented instance the network predicts the type of nucleus via a devoted up-sampling branch. We demonstrate state-of-the-art performance compared to other methods on multiple independent multi-tissue histology image datasets. As part of this work, we introduce a new dataset of Haematoxylin \& Eosin stained colorectal adenocarcinoma image tiles, containing 24,319 exhaustively annotated nuclei with associated class labels.  
\end{abstract}

\begin{IEEEkeywords}
		Nuclear segmentation, nuclear classification, computational pathology, deep learning.
	\end{IEEEkeywords}

\IEEEpeerreviewmaketitle


\section{Introduction} \label{section:intro}
 Current manual assessment of Haematoxylin and Eosin (H\&E) stained histology slides suffers from low throughput and is naturally prone to intra- and inter-observer variability  \cite{elmore2015diagnostic}. To overcome the difficulty in visual assessment of tissue slides, there is a growing interest in digital pathology (DP), where digitised whole-slide images (WSIs) are acquired from glass histology slides using a scanning device. This permits efficient processing, analysis and management of the tissue specimens \cite{madabhushi2016digital_pathology}. Each WSI contains tens of thousands of nuclei of various types, which can be further analysed in a systematic manner and used for predicting clinical outcome. Here, the type of nucleus refers to the cell type in which it is located. For example, nuclear features can be used to predict survival~\cite{alsubaie2018bottom} and also for diagnosing the grade and type of disease \cite{lu2018nuclear}. Also, efficient and accurate detection and segmentation of nuclei can facilitate good quality tissue segmentation \cite{sirinukunwattana2018novel, javed2018cellular}, which can in turn not only facilitate the quantification of WSIs but may also serve as an important step in understanding how each tissue component contributes to disease. In order to use nuclear features for downstream analysis within computational pathology, nuclear segmentation must be carried out as an initial step. However, this remains a challenge because nuclei display a high level of heterogeneity and there is significant inter- and intra-instance variability in the shape, size and chromatin pattern between and within different cell types, disease types or even from one region to another within a single tissue sample. Tumour nuclei, in particular, tend to be present in clusters, which gives rise to many overlapping instances, providing a further challenge for automated segmentation, due to the difficulty of separating neighbouring instances. 

    As well as extracting each individual nucleus, determining the type of each nucleus can increase the diagnostic potential of current DP pipelines. For example, accurately classifying each nucleus to be from tumour or lymphocyte enables downstream analysis of tumour infiltrating lymphocytes (TILs), which have been shown to be predictive of cancer recurrence \cite{corredor2019spatial}. Yet, similar to nuclear segmentation, classifying the type of each nucleus is difficult, due to the high variance of nuclear appearance within each WSI. Typically, nuclei are classified using two disjoint models: one for detecting each nucleus and then another for performing nuclear classification  \cite{sharma2015multi, wang2016automatic}. However, it would be preferable to utilise a single unified model for nuclear instance segmentation and classification.
    
    In this paper, we present a deep learning approach\footnote{Model code: \url{https://github.com/vqdang/hover\_net}} for simultaneous segmentation and classification of nuclear instances in histology images. The network is based on the prediction of horizontal and vertical distances (and hence the name HoVer-Net) of nuclear pixels to their centres of mass, which are subsequently leveraged to separate clustered nuclei. For each segmented instance, the nuclear type is subsequently determined via a dedicated up-sampling branch. To the best of our knowledge, this is the first approach that achieves instance segmentation and classification within the same network. We present comparative results on six independent multi-tissue histology image datasets and demonstrate state-of-the-art performance compared to other recently proposed methods. The main contributions of this work are listed as follows:
	
	\begin{itemize}
	    
	    \item A novel network, targeted at simultaneous segmentation and classification of nuclei, where horizontal and vertical distance map predictions separate clustered nuclei.
	    \item We show that the proposed HoVer-Net achieves state-of-the-art performance on multiple H\&E histology image datasets, as compared to over a dozen recently published methods. 
	    \item An interpretable and reliable evaluation framework that effectively quantifies nuclear segmentation performance and overcomes the limitations of existing performance measures.
	    \item A new dataset\footnote{The CoNSeP dataset for nuclear segmentation is available at \url{https://warwick.ac.uk/fac/sci/dcs/research/tia/data/}.} of 24,319 exhaustively annotated nuclei within 41 colorectal adenocarcinoma image tiles. 
	\end{itemize}
	
	\section{Related Work}
	\subsection{Nuclear Instance Segmentation}
	
	Within the current literature, \textbf{energy-based} methods, in particular the watershed algorithm, have been widely utilised to segment nuclear instances. For example, \cite{yang2006nuclei} used thresholding to obtain the markers and the energy landscape as input for watershed to extract the nuclear instances. Nonetheless, thresholding relies on a consistent difference in intensity between the nuclei and background, which does not hold for more complex images and hence often produces unreliable results. Various approaches have tried to provide an improved marker for marker-controlled watershed. \cite{cheng2009segmentation} used active contours to obtain the markers. \cite{veta2013} used a series of morphological operations to generate the energy landscape. However, these methods rely on the predefined geometry of the nuclei to generate the markers, which determines the overall accuracy of each method. Notably, \cite{ali2012integrated} avoided the trouble of refining the markers for watershed by designing a method that relies solely on the energy landscape. They combined an active contour approach with nuclear shape modelling via a level-set method to obtain the nuclear instances. Despite its widespread usage, obtaining sufficiently strong markers for watershed is a non-trivial task. Some methods have departed from the energy-based approach by utilising the geometry of the nuclei. For instance, \cite{wienert2012detection}, \cite{latorre2013concave} and \cite{kwak2015nucleus} computed the concavity of nuclear clusters, while \cite{liao2016eclipse} used eclipse-fitting to separate the clusters. However, this assumes a predefined shape, which does not encompass the natural diversity of the nuclei. In addition, these methods tend to be sensitive to the choice of manually selected parameters.
	
	Recently, \textbf{deep learning} methods have received a surge of interest due to their superior performance in many computer vision tasks \cite{litjens2017survey,shen2017deep,lecun2015deep}. These approaches are capable of automatically extracting a representative set of features, that strongly correlate with the task at hand. As a result, they are preferable to hand-crafted approaches, that rely on a selection of pre-defined features. Inspired by the Fully Convolutional Network (FCN) \cite{long2015fully}, U-Net \cite{ronneberger2015u} has been successfully applied to numerous segmentation tasks in medical image analysis. The network has an encoder-decoder design with skip connections to incorporate low-level information and uses a \textbf{weighted loss function} to assist separation of instances. However, it often struggles to split neighbouring instances and is highly sensitive to pre-defined parameters in the weighted loss function. A more recently proposed method in Micro-Net \cite{micronet2018Shan} extends U-Net by utilising an enhanced network architecture with weighted loss. The network processes the input at multiple resolutions and as a result, gains robustness against nuclei with varying size. In \cite{graham2018b}, the authors developed a network that is robust to stain variations in H\&E images by introducing a weighted loss function that is sensitive to the Haematoxylin intensity within the image. 
	
	Other methods exploit information about the nuclear \textbf{contour} (or boundary) within the network, such as DCAN \cite{chen2016dcan} that utilised a dual architecture that outputs the nuclear cluster and the nuclear contour as two separate prediction maps. Instance segmentation is then achieved by subtracting the contour from the nuclear cluster prediction. \cite{cui2018deep} proposed a network to predict the inner nuclear instance, the nuclear contour and the background. The network utilised a customised weighted loss function based on the relative position of pixels within the image to improve and stabilise the inner nuclei and contour prediction. Some other methods have also utilised the nuclear contour to achieve instance segmentation. For example, \cite{kumar} employed a deep learning technique for labelling the nuclei and the contours, followed by a region growing approach to extract the final instances. \cite{khoshdeli2018deep} used the contour predictions as input into a further network for segmentation refinement. \cite{zhou2019cia} proposed CIA-Net, that utilises a multi-level information aggregation module between two task-specific decoders, where each decoder segments either the nuclei or the contours. A Deep Residual Aggregation Network (DRAN) was proposed by \cite{vu2018methods} that uses a multi-scale strategy, incorporating both the nuclei and nuclear contours to accurately segment nuclei.
	
	There have been various other methods to achieve instance separation. Instead of considering the contour, \cite{naylor2018segmentation} proposed a deep learning approach to detect superior markers for watershed by regressing the nuclear \textbf{distance map}. Therefore, the network avoids making a prediction for areas with indistinct contours.
	
	In line with these developments, the field of instance segmentation within natural images is also rapidly progressing and have had a significant influence on nuclear instance segmentation methods. A notable example is Mask-RCNN \cite{mrcnn}, where instance segmentation approach is achieved by first predicting candidate regions likely to contain an object and then deep learning based segmentation within those proposed regions.
	
		\begin{figure*}[t!]
		\centering
		\includegraphics[width=1.0\textwidth]{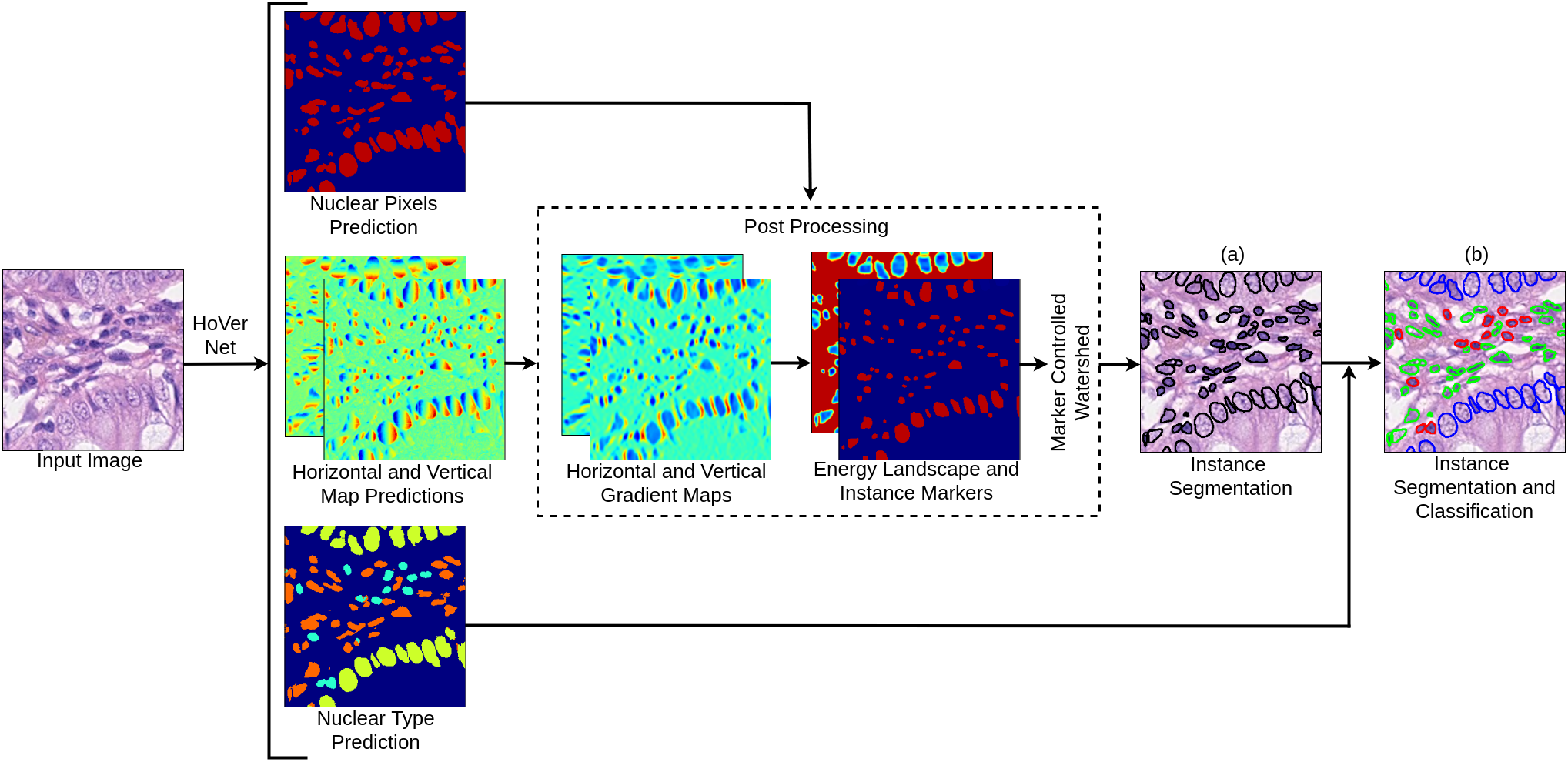}
		\caption{Overview of the proposed approach for simultaneous nuclear instance segmentation and classification. When no classification labels are available, the network produces the instance segmentation as shown in (a). The different colours of the nuclear boundaries represent different types of nuclei in (b).}
		\label{fig:pipeline}
	\end{figure*}
	
	\subsection{Nuclear Classification}
	As well as performing instance segmentation, it is desirable to determine the \textit{type} of each nucleus to facilitate and improve downstream analysis. It is possible for current models to differentiate between certain nuclear types in H\&E, however sub-typing of lymphocytes is an extremely hard task due to the high levels of similarity in morphological appearance between T and B lymphocytes. Typically, classifying each nucleus is done via a two-stage approach, where the first step involves either nuclear segmentation or nuclear detection. When segmentation is used as the initial step, a series of morphological and textural features are extracted from each instance, which are then used within a classifier to determine the nuclei classes. For example, \cite{nguyen2011prostate} classified nuclei within H\&E stained breast cancer images as either tumour, lymphocyte or stromal based on their morphological features. \cite{yuan2012quantitative} performed nuclear segmentation and then classified each nucleus with AdaBoost classifier, utilising the intensity, morphology and texture of nuclei as features. Otherwise, detection is performed as an initial step and a patch centred at the point of detection is fed into a classifier, to predict the type of nucleus. \cite{sirinukunwattana2016locality} proposed a spatially constrained CNN, that initially detects all nuclei and then for each nucleus an ensemble of associated patches are fed into a CNN to predict the type to be either epithelial, inflammatory, fibroblast or miscellaneous.

	%%%%%%%%%%%%%%%%%%%%%%%%%%%%%%%%%%%%%%%%%%%%%%%%%%%%%%%%%%%%%%%%%%%%%%%%%%%%%%%%%%%%%%%%%%%%
	
	\section{Methods} \label{section:methods}
\begin{figure*}[!t]
		\centering
		\includegraphics[width=1.0\textwidth]{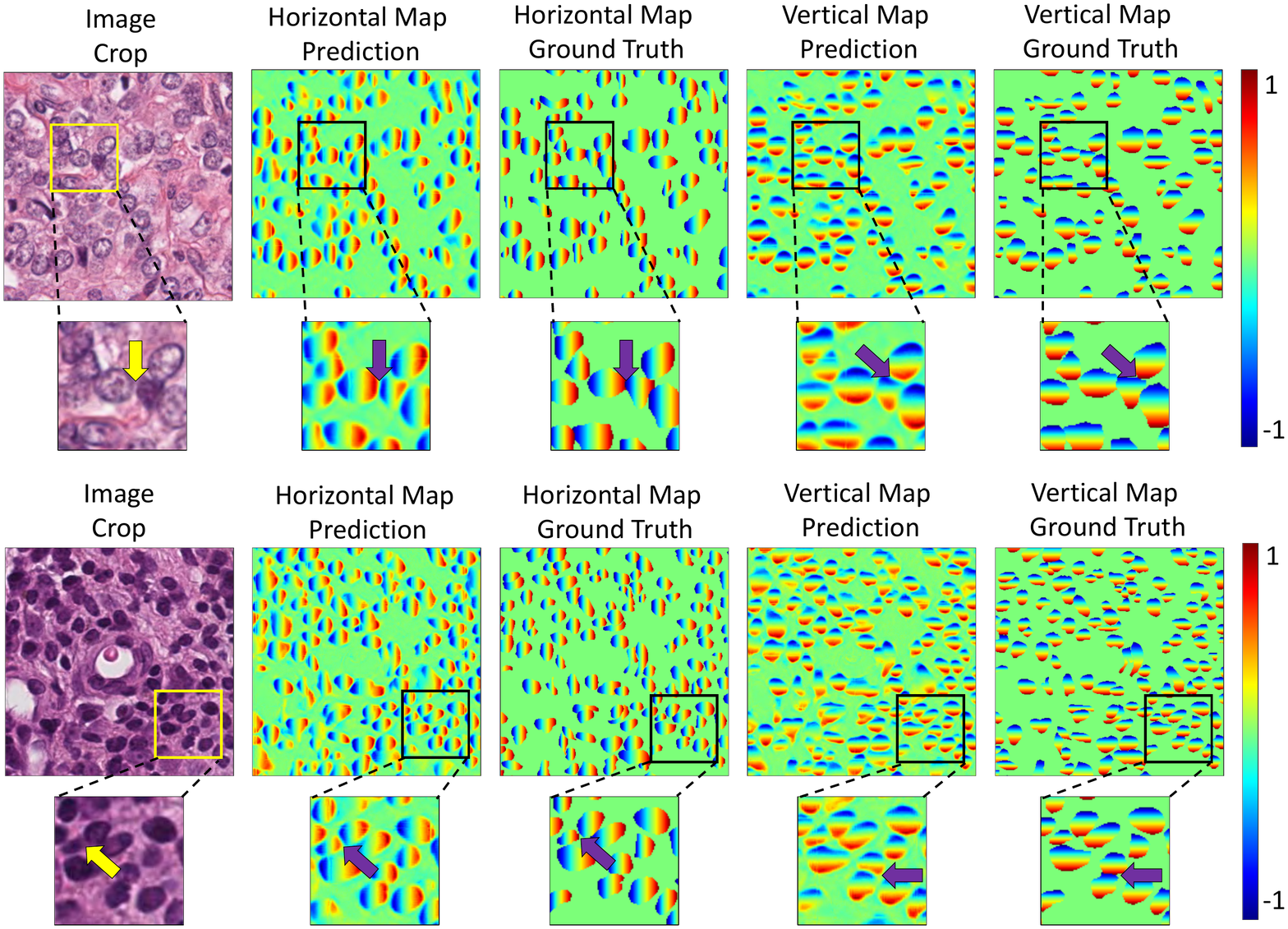}
		\caption{Cropped image regions showing horizontal and vertical map predictions, with corresponding ground truth. Arrows highlight the strong instance information encoded within these maps, where there is a significant difference in the pixel values.}
		\label{fig:xy_output}
	\end{figure*}

	Our overall framework for automatic nuclear instance segmentation and classification can be observed in Fig. \ref{fig:pipeline} and the proposed network in Fig. \ref{fig:xy_network}. Here, nuclear pixels are first detected and then, a tailored post-processing pipeline is used to simultaneously segment nuclear instances and obtain the corresponding nuclear types. The framework is based upon the horizontal and vertical distance maps, which can be seen in Fig. \ref{fig:xy_output}. In the figure, each nuclear pixel denotes either the horizontal or vertical distance of pixels to their centres of mass.
	
	\subsection{Network Architecture} \label{section:networkarchitecture}
    In order to extract a strong and representative set of features, we employ a deep neural network. The feature extraction component of the network is inspired by the pre-activated residual network with 50 layers \cite{preact_resnet} (Preact-ResNet50), due to its excellent performance in recent computer vision tasks \cite{imagenet_cvpr09} and robustness against input perturbation \cite{anurag2018resnet_robustness}. Compared to the standard Preact-ResNet50 implementation, we reduce the total down-sampling factor from 32 to 8 by using a stride of 1 in the first convolution and removing the subsequent max-pooling operation. This ensures that there is no immediate loss of information that is important for performing an accurate segmentation. Various residual units are applied throughout the network at different down-sampling levels. A series of consecutive residual units is denoted as a residual block. The number of residual units within each residual block is 3, 4, 6 and 3 that are applied at down-sampling levels 1, 2, 4 and 8 respectively. For clarity, a down-sampling level of 2 means that the input has a reduction in the spatial resolution by a factor of 2.

	\begin{figure*}[t!]
		\centering
		\includegraphics[width=1.0\textwidth]{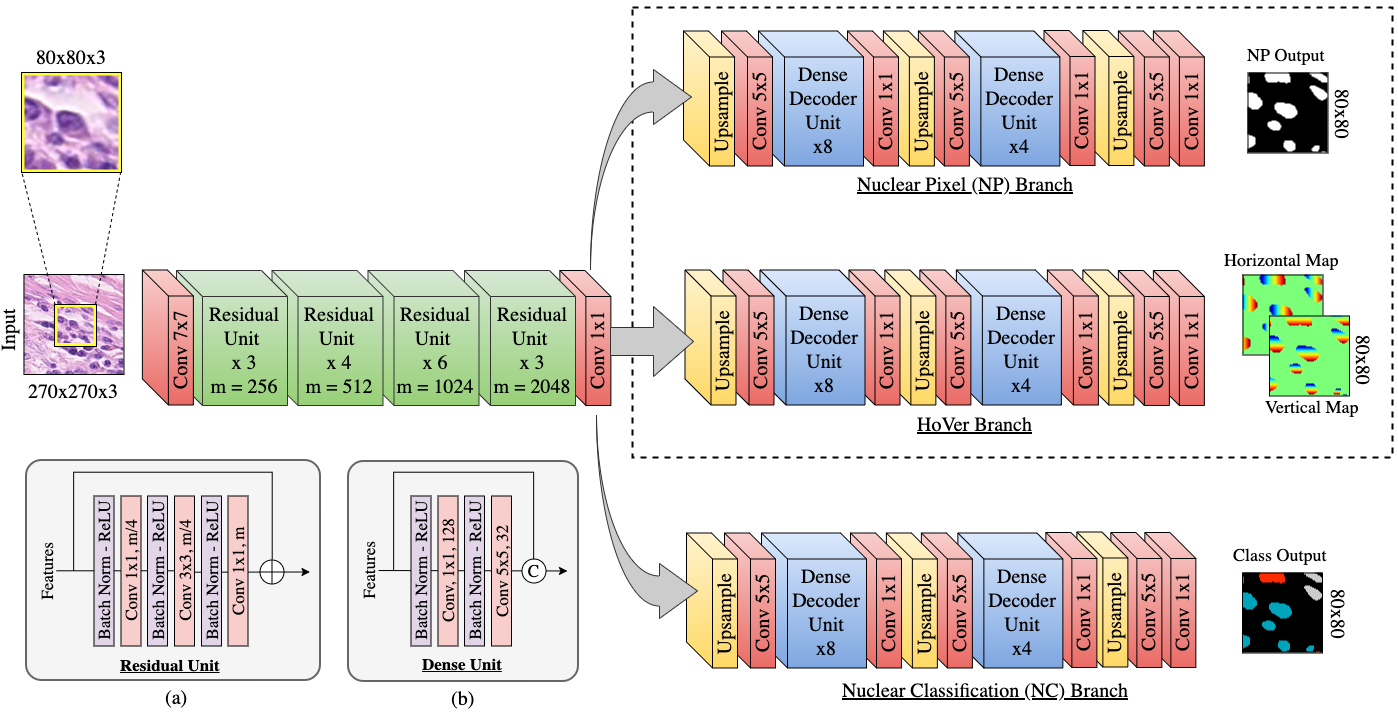}
		\caption{Overview of the proposed architecture. (a) (Pre-activated) residual unit, (b) dense unit. $m$ indicates the number of feature maps within each residual unit. The yellow square within the input denotes the considered region at the output. When the classification labels aren't available, only the up-sampling branches in the dashed box are considered.}
		\label{fig:xy_network}
	\end{figure*}

    Following Preact-ResNet50, we perform nearest neighbour up-sampling via three distinct branches to simultaneously obtain accurate nuclear instance segmentation and classification. We name the corresponding branches: (i) nuclear pixel (NP) branch; (ii) HoVer branch and (iii) nuclear classification (NC) branch. The NP branch predicts whether or not a pixel belongs to the nuclei or background, whereas the HoVer branch predicts the horizontal and vertical distances of nuclear pixels to their centres of mass. Then, the NC branch predicts the type of nucleus for each pixel. In particular, the NP and HoVer branches jointly achieve nuclear instance segmentation by first separating nuclear pixels from the background (NP branch) and then separating touching nuclei (HoVer branch). The NC branch determines the type of each nucleus by aggregating the pixel-level nuclear type predictions within each instance.
    
    All three up-sampling branches utilise the same architectural design, which consists of a series of up-sampling operations and densely connected units \cite{densenet} (or dense units). By stacking multiple and relatively cheap dense units, we build a large receptive field with minimal parameters, compared to using a single convolution with a larger kernel size and we ensure efficient gradient propagation. We use skip connections \cite{ronneberger2015u} to incorporate features from the encoder, but utilise summation as opposed to concatenation. The consideration of low-level information is particularly important in segmentation tasks, where we aim to precisely delineate the object boundaries. We use dense units after the first and second up-sampling operations, where the number of units is 4 and 8 respectively. Valid convolution is performed throughout the two up-sampling branches to prevent poor predictions at the boundary. This results in the size of the output being smaller than the size of the input. As opposed to using a dedicated network for each task, a shared encoder makes it possible to train the nuclear instance segmentation and classification model end-to-end and therefore, reduce the total training time. Furthermore, a shared encoder can also take advantage of the shared information across multiple tasks and thus, help to improve the model performance on all tasks. 
    
    Finally, if we do not have the classification labels of the nuclei, only the NP and HoVer up-sampling branches are considered. Otherwise, we consider all three up-sampling branches and perform simultaneous nuclear instance segmentation and classification. 
    
    We display an overview of the network architecture in Fig. \ref{fig:xy_network}, where the spatial dimension of the input is 270$\times$270 and the output dimension of each branch is 80$\times$80. The dashed box within Fig. \ref{fig:xy_network} highlights the branches for nuclear instance segmentation. Additionally, we also show a residual unit and a dense unit within Fig. \ref{fig:xy_network}a and Fig. \ref{fig:xy_network}b. We denote $m$ as the number of feature maps within each convolution of a given residual unit. At each down sampling level, from left to right, $m$=256, 512, 1024, 2048 respectively. We keep a fixed amount of feature maps within each dense unit throughout the two branches as shown in Fig. \ref{fig:xy_network}c.

	\subsubsection{Loss Function} \label{section:loss}
	
	The proposed network design has 4 different sets of weights: $w_0$, $w_1$,  $w_2$ and $w_3$ which refer to the weights of the Preact-ResNet50 encoder, the HoVer branch decoder, the NP branch decoder and the NC branch decoder. These 4 sets of weights are optimised jointly using the loss $\mathcal{L}$ defined as:
	% we dont have regularization terms
	\begin{equation}
    	\small
    	\mathcal{L} = 
    	    \underbrace{\lambda_a\mathcal{L}_{a} + \lambda_b\mathcal{L}_{b}}_{\text{HoVer Branch}}
	      + \underbrace{\lambda_c\mathcal{L}_{c} + \lambda_d\mathcal{L}_{d}}_{\text{NP Branch}}
	      + \underbrace{\lambda_e\mathcal{L}_{e} + \lambda_f\mathcal{L}_{f}}_{\text{NC Branch}}
	\end{equation}
	\noindent where $\mathcal{L}_{a}$ and $\mathcal{L}_{b}$ represent the regression loss with respect to the output of the HoVer branch, $\mathcal{L}_{c}$ and $\mathcal{L}_{d}$ represent the loss with respect to the output at the NP branch and and finally, $\mathcal{L}_{e}$ and $\mathcal{L}_{f}$ represent the loss with respect to the output at the NC branch. We choose to use two different loss functions at the output of each branch for an overall superior performance. $\lambda_a ... \lambda_f$ are scalars that give weight to each associated loss function. Specifically, we set $\lambda_b$ to 2 and the other scalars to 1, based on empirical selection.

	Given the input image $I$, at each pixel $i$ we define $p_i(I, w_0, w_1)$ as the regression output of the HoVer branch, whereas $q_i(I, w_0, w_2)$ and $r_i(I, w_0, w_3)$ denote the pixel-based softmax predictions of the NP and NC branches respectively. We also define $\Gamma_i(I)$, $\Psi_i(I)$ and $\Phi_i(I)$ as their corresponding ground truth (GT). $\Psi_i(I)$ is the GT of the nuclear binary map, where background pixels have the value of 0 and nuclear pixels have the value 1. On the other hand, $\Phi_i(I)$ is the nuclear type GT where background pixels have the value 0 and any integer value larger than 0 indicates the type of nucleus. Meanwhile, $\Gamma_i(I)$ denotes the GT of the horizontal and vertical distances of nuclear pixels to their corresponding centres of mass. For $\Gamma_i(I)$, we assign values between -1 and 1 to nuclear pixels in both the horizontal and vertical directions. We assign the value of the background and the line crossing the centre of mass within each nucleus to be 0. For clarity, we denote the horizontal and vertical components of the GT HoVer map as horizontal map $\Gamma _{i,x}$ and vertical map $\Gamma _{i,y}$ respectively. Visual examples of the horizontal and vertical maps can be seen in Fig. \ref{fig:xy_output}.
	
	At the output of the HoVer branch, we compute a multiple term regression loss. We denote $\mathcal{L}_a$ as the mean squared error between the predicted horizontal and vertical distances and the GT. We also propose a novel loss function $\mathcal{L}_b$ that calculates the mean squared error between the horizontal and vertical gradients of the horizontal and vertical maps respectively and the corresponding gradients of the GT. We formally define $\mathcal{L}_a$ and $\mathcal{L}_b$ as:
	\begin{equation}
	\small
	\mathcal{L}_a = \frac{1}{n}\sum_{i = 1}^n{(p_i(I;\mathbold{w}_0,\mathbold{w}_1)- \Gamma _i(I))^2} 
	\end{equation}
	\begin{equation}
	\small
	\begin{split}
	\mathcal{L}_b =
	\frac{1}{m}\sum_{i \in M}{(\nabla_x(p_{i,x}(I;\mathbold{w}_0,\mathbold{w}_1))-\nabla_x( \Gamma _{i,x}(I)))^2} 
	\\ +
	\frac{1}{m}\sum_{i \in M}{(\nabla_y(p_{i,y}(I;\mathbold{w}_0, \mathbold{w}_1))-\nabla_y( \Gamma _{i,y}(I)))^2}
	\end{split}
	\end{equation}
	\noindent Within equation (3), $\nabla_x$ and $\nabla_y$ denote the gradient in the horizontal $x$ and vertical $y$ directions respectively. $m$ denotes total number of nuclear pixels within the image and $M$ denotes the set containing all nuclear pixels.
	
	At the output of NP and NC branches, we calculate the cross-entropy loss ($\mathcal{L}_{c}$ and $\mathcal{L}_{e}$) and the dice loss ($\mathcal{L}_{d}$ and $\mathcal{L}_{f}$). These two losses are then added together to give the overall loss of each branch. Concretely, we define the cross entropy and dice losses as:
	\begin{equation}
	\small
	\begin{split}
	\text{CE} = -\frac{1}{n}\sum_{i=1}^N
	    \sum_{k=1}^K
	    X_{i,k}(I)\log Y_{i,k}(I)
	\end{split}
	\end{equation}
	\begin{equation}
	\small
	\text{Dice} = 1- \frac{2 \times  \sum_{i=1}^N (Y_i(I) \times X_i(I))+ \epsilon}{\sum_{i=1}^N{Y_i(I) +\sum_{i=1}^N X _i(I)} + \epsilon}
	\end{equation}
	\noindent where $X$ is the ground truth, $Y$ is the prediction, $K$ is the number of classes and $\epsilon$ is a smoothness constant which we set to $1.0e^{-3}$. When calculating $\mathcal{L}_{c}$ and $\mathcal{L}_{d}$ for NP branch, for a given pixel $i$, we set $X_i$ and $Y_i$ as $q_i(I, w_0, w_2)$ and $\Psi_i$ respectively. For $\mathcal{L}_{c}$, we set $K$ to be 2 within equation (4) because the task of the branch is to perform binary nuclear segmentation. Similarly, for $\mathcal{L}_{e}$ and $\mathcal{L}_{f}$ at NC branch, for a given pixel $i$, we substitute $X_i$ for $\Phi_i(I)$ and $Y_i$ for $r_i(I, w_0, w_3)$ in equations (4) and (5). $K$ is set as 5 within equation (4) when calculating $\mathcal{L}_{e}$, denoting the 4 types of nuclei that our model currently predicts and the background. Note, the value of $K$ is chosen to reflect the number of nuclear types represented in the training set. 
	
	It must be noted that the NC branch loss $\mathcal{L}_{e}$ and $\mathcal{L}_{f}$ are only calculated when the classification labels are available. In other words, as mentioned in Section \ref{section:networkarchitecture}, the network performs only instance segmentation if there are no classification labels given.

    \subsection{Post Processing}  \label{section:postprocessing}
	% need major editing
	Within each horizontal and vertical map, pixels between separate instances have a significant difference. This can be seen in Fig. \ref{fig:xy_output} and is highlighted by the arrows. Therefore, calculating the gradient can inform where the nuclei should be separated because the output will give high values between neighbouring nuclei, where there is a significant difference in the pixel values. We define:
	\begin{equation}
	\begin{split}
	\small
	\mathcal{S}_m = max(H_x(p_{x}), H_y(p_{y}))
	\end{split}
	\end{equation}
	where $p_x$ and $p_y$ refer to the the horizontal and vertical predictions at the output of the HoVer branch and $H_x$ and $H_y$ refer to the horizontal and vertical components of the Sobel operator. Specifically, $H_x$ and $H_y$ compute the horizontal and vertical derivative approximations and are shown by the gradient maps in Fig. \ref{fig:pipeline}. Therefore, $\mathcal{S}_m$ highlights areas where there is a significant difference in neighbouring pixels within the horizontal and vertical maps. Therefore, areas such as the ones shown by the arrows in Fig. \ref{fig:xy_output} will result in high values within $\mathcal{S}_m$. We compute markers $M$ = $\sigma(\tau(q, h) - \tau(S_m, k))$. Here, $\tau(a, b)$ is a threshold function that acts on $a$ and sets values above $b$ to 1 or 0 otherwise. Specifically, $h$ and $k$ were chosen such that they gave the optimal nuclear segmentation results. $\sigma$ is a rectifier that sets all negative values to 0 and $q$ is the probability map output of the NP branch. We obtain the energy landscape $E=[1-\tau(S_m, k)]*\tau(q, h)$. Finally, $M$ is used as the marker during marker-controlled watershed to determine how to split $\tau(q, h)$, given the energy landscape $E$. This sequence of events can be seen in Fig. \ref{fig:pipeline}.
	
			\begin{figure}[!t]
		\centering
        \includegraphics[width=1.0\columnwidth]{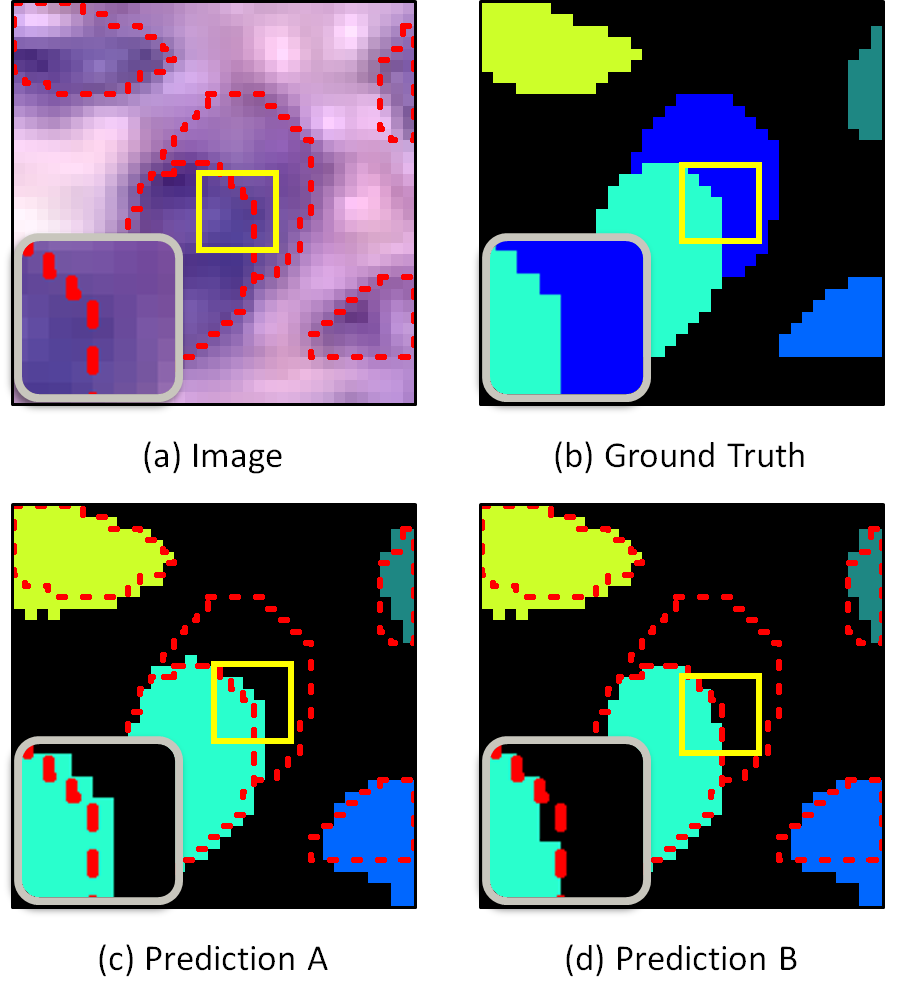}
		\caption{Examples highlighting the limitations of DICE2 and AJI with slightly different predictions. For better visualisation, ground truth contours (red dash line) for each instance have been overlaid on both the predictions and original images.} 
		\label{fig:dice2_aji_problem}
	\end{figure}

	To perform simultaneous nuclear instance segmentation and classification, it is necessary to convert the per-pixel nuclear type prediction at the output of the NC branch to a prediction per nuclear instance. For each nuclear instance, we use {\em majority class} of the predictions made by the NC branch, i.e., the nuclear type of all pixels in an instance is assigned to be the class with the highest frequency count for that nuclear instance.
	
	   \begin{table}[!t]
	\begin{center}
		\caption{Comparison between Prediction \textit{A} and Prediction \textit{B} from Fig.\ref{fig:dice2_aji_problem} across various measurements.}
		\setlength{\tabcolsep}{6pt} % Default value: 6pt
		\renewcommand{\arraystretch}{1.0} % Default value: 1
		\label{table:metrics_toy_results}
		\begin{tabular}{c|cccc}
			& \textbf{DICE2} & \textbf{AJI} & \textbf{PQ} \\
			\midrule
            Prediction \textit{A} & 0.6477 & 0.4790 & 0.6803 \\
            Prediction \textit{B} & 0.9007 & 0.6414 & 0.6863 \\
			\bottomrule
		\end{tabular}
	\end{center}
	\end{table}

	Please refer to Appendix A for a full analysis on the contribution of our proposed loss function, post-processing method and devoted classification branch.
	
	\section{Evaluation Metrics} \label{section:metrics}
	% instance segmentation, instance must come first, quality of pixel-wise of each instance come after !!! need to rewrite to stress this
    \subsection{Nuclear Instance Segmentation Evaluation} \label{section:metrics_inst}

    Assessment and comparison of different methods is usually given by an overall score that indicates which method is superior. However, to further investigate the method, it is preferable to break the problem into sub-tasks and measure the performance of the method on each sub-task. This enables an in depth analysis, thus facilitating a comprehensive understanding of the approach, which can help drive forward model development. For nuclear instance segmentation, the problem can be divided into the following three sub-tasks:
	\begin{itemize}
		\item Separate the nuclei from the background
		\item Detect individual nuclear instances
		\item Segment each detected instance
	\end{itemize}
	In the current literature, two evaluation metrics have been mainly adopted to quantitatively measure the performance of nuclear instance segmentation: 1) Ensemble Dice (DICE2) \cite{vu2018methods}, and 2) Aggregated Jaccard Index (AJI) \cite{kumar}. Given the ground truth $X$ and prediction $Y$, DICE2 computes and aggregates DICE per nucleus, where Dice coefficient (DICE) is defined as 2$\times$($X$$\cap$$Y$)$\slash$($\mid$$X$$\mid$+$\mid$$Y$$\mid$) and AJI computes the ratio of an aggregated intersection cardinality and an aggregated union cardinality between $X$ and $Y$. 
	
	These two evaluation metrics only provide an overall score for the instance segmentation quality and therefore provides no further insight into the sub-tasks at hand. In addition, these two metrics have a limitation, which we illustrate in Fig. \ref{fig:dice2_aji_problem}. From the figure, although prediction \textit{A} only differs from prediction \textit{B} by a few pixels, the DICE2 and AJI scores for \textit{B} are inferior. These scores are shown in Table \ref{table:metrics_toy_results}. This problem arises due to over-penalisation of the overlapping regions. By overlaying the GT segment contours (red dashed line) upon the two predictions, we observe that, although the cyan-coloured instance within prediction \textit{A} overlaps mostly with the cyan-coloured GT instance, it also slightly overlaps with the blue-coloured GT instance. As a result, according to the DICE2 algorithm, the predicted cyan instance will be penalised by pixels not only coming from the dominant overlapping cyan-coloured GT instance, but also from the blue-coloured GT instance. The AJI also suffers from the same phenomenon. However, because AJI only uses the prediction and GT instance pair with the highest intersection over union, over-penalisation is less likely compared to DICE2. Over-penalisation is likely to occur when the model completely fails to detect the neighbouring instance, such as in Fig. \ref{fig:dice2_aji_problem}. Nonetheless, when evaluating methods across different datasets, specifically on samples containing lots of hard to recognise nuclei such as fibroblasts or nuclei with poor staining, the number of failed detections may increase and therefore may have a negative impact on the AJI measurement. Due to the limitations of DICE2 and AJI, it is clear that there is a need for an improved reliable quantitative measurement.
	
	\textbf{Panoptic Quality}: We propose to use another metric for accurate quantification and interpretability to assess the performance of nuclear instance segmentation. Originally proposed by \cite{kirilov2018panoptic_quality}, panoptic quality (PQ) for nuclear instance segmentation is defined as:
	\begin{equation}
	\small
	\mathcal{PQ}= 
	\underbrace{\frac{|TP|}{|TP|+\frac{1}{2}|FP|+\frac{1}{2}|FN|}}_{\text{Detection Quality(DQ)}}
	\times
	\underbrace{{\frac{\sum_{(x,y)\in{TP}}{IoU(x,y)}}{|TP|}}}_{\text{Segmentation Quality(SQ)}}
	\end{equation}
	where \textit{x} denotes a GT segment, \textit{y} denotes a prediction segment and IoU denotes intersection over union. Each (\textit{x,y}) pair is mathematically proven to be \textit{unique} \cite{kirilov2018panoptic_quality} over the entire set of prediction and GT segments if their IoU(\textit{x,y})$>$0.5. The unique matching splits all available segments into matched pairs (TP), unmatched GT segments (FN) and unmatched prediction segments (FP). From this, PQ can be intuitively analysed as follows: the \textit{detection quality} (DQ) is the F$_1$ Score that is widely used to evaluate instance detection, while \textit{segmentation quality} (SQ) can be interpreted as how close each correctly detected instance is to their matched GT. DQ and SQ, in a way, also provide a direct insight into the second and third sub-tasks, defined above. We believe that PQ should set the standard for measuring the performance of nuclear instance segmentation methods.
	
    Overall, to fully characterise and understand the performance of each method, we use the following three metrics: 1) DICE to measure the separation of all nuclei from the background; 2) Panoptic Quality as a unified score for comparison and 3) AJI for direct comparison with previous publications\footnote{Evaluation code:  \url{https://github.com/vqdang/hover\_net/src/metrics}}. Panoptic quality is further broken down into DQ and SQ components for interpretability. Note, SQ is calculated only within true positive segments and should therefore be observed together with DQ. Throughout this study, these metrics are calculated for each image and the average of all images are reported as final values for each dataset.
    \subsection{Nuclear Classification Evaluation} \label{section:metrics_class}
   
   Classification of the type of each nucleus is performed within the nuclear instances extracted from the instance segmentation or detection tasks. Therefore, the overall measurement for nuclear type classification should also encompass these two tasks. For all nuclear instances of a particular type $t$ from both the ground truth and the prediction, the detection task $d$ splits the GT and predicted instances into the following subsets: correctly detected instances (TP$_d$), misdetected GT instances (FN$_d$) and overdetected predicted instances (FP$_d$). Subsequently, the classification task $c$ further breaks TP$_d$ into correctly classified instances of type $t$ (TP$_c$), correctly classified instances of types other than type $t$ (TN$_c$), incorrectly classified instances of type $t$ (FP$_c$) and incorrectly classified instances of types other than type $t$ (FN$_c$). We then define the F$_c$ score of each type $t$ for combined nuclear type classification and detection as follows:
    \begin{equation}
	\small
       F_c^t = \frac
                {2 (TP_c + TN_c)}
                {\Bigg[\begin{matrix}2 (TP_c + TN_c) + \alpha_0 FP_c + \alpha_1 FN_c&\\
                + \alpha_2 FP_d + \alpha_3 FN_d\end{matrix}\Bigg]}
    \end{equation}
    \noindent where we use $\alpha_0=\alpha_1=2$ and $\alpha_2=\alpha_3=1$ to give more emphasis to nuclear type classification. Moreover, using the same weighting, if we further extend $t$ to encompass all types of nuclei $T$ ($t \in T$), the classification within TP$_d$ is then divided into a correctly classified set $A_{c}$ and an incorrectly classified set $B_{c}$. We can therefore disassemble $F_c^t$ into:
    \begin{equation}
	\small
	\begin{split}
        F_c^T
              & = \frac{2 A_c}{2 (A_{c} + B_{c}) + FP_d + FN_d} \\
              & = \frac {2 (A_{c} + B_{c})}{2 (A_{c} + B_{c}) + FP_d + FN_d} \times \frac{A_{c}}{A_{c} + B_{c}} \\
              & = F_d \times \begin{matrix}\textnormal{Classification Accuracy within}&\\\textnormal{Correctly Detected Instances}\end{matrix}
    \end{split}
    \end{equation}
    where $F_d$ is simply the standard detection quality like DQ while the other term is the accuracy of nuclear type classification within correctly detected instances. In the case where the GT is not exhaustively annotated for nuclear type classification, like in CRCHisto, an amount equal to the number of unlabelled GT instances in each set is subtracted from $B_c$ and $FN_c$. 
    
    Finally, while IoU is utilised as the criteria in DQ for selecting the TP for detection in instance segmentation, detection methods can not calculate the IoU. Therefore, to facilitate comparison of both instance segmentation and detection methods for the nuclear type classification tasks, for $F_c^t$, we utilise the notion of distance to determine whether nuclei have been detected. To be precise, we define the region within a predefined radius from the annotated centre of the nucleus as the ground truth and if a prediction lies within this area, then it is considered to be a true positive. Here, we are consistent with \cite{sirinukunwattana2016locality} and use a radius of 6 pixels at 20$\times$ or 12 pixels at 40$\times$. 
    
	\section{Experimental Results} \label{section:expandresults}
	\subsection{Datasets} \label{section:datasets}
	
	% kumar-train 9796, kumar-st 3140, kumar-dt 4024
	% cpm-17, train 3803, test 3736
	% cpm-15 2936
	% tnbc 4056
	
	\begin{table*}[!t]
    	\begin{center}
    		\caption{Summary of the datasets used in our experiments. UHCW denotes University Hospitals Conventry and Warwickshire and TCGA denotes The Cancer Genome Atlas. {\em Seg} denotes segmentation masks and {\em Class} denotes classification labels.}
    		\label{table:dataset_summary}
    		\setlength{\tabcolsep}{9pt} % Default value: 6pt
    		\renewcommand{\arraystretch}{1} % Default value: 1
    		\resizebox{\textwidth}{!}{
    			\begin{tabular}{c|cccccc}
    				& \textbf{CoNSeP} & \textbf{Kumar} & \textbf{CPM-15} & \textbf{CPM-17}& \textbf{TNBC} & \textbf{CRCHisto} \\
    				\toprule
    				Total Number of Nuclei     & 24,319 & 21,623 & 2,905 & 7,570 & 4,056  & 29,756  \\
    				Labelled Nuclei & 24,319 & 0 & 0 & 0 & 0  & 22,444  \\
    				Number of Images & 41 & 30 & 15 & 32 & 50  & 100  \\
    				Origin     & UHCW & TCGA & TCGA & TCGA & Curie Institute  & UHCW  \\
    				Magnification    & 40$\times$ & 40$\times$ & 40$\times$ \& 20$\times$ & 40$\times$ \& 20$\times$ & 40$\times$ & 20$\times$  \\
    				Size of Images     & 1000$\times$1000 & 1000$\times$1000 & 400$\times$400 to 1000$\times$600 & 500$\times$500 to 600$\times$600 & 512$\times$512 & 500$\times$500  \\
    				{\em Seg/Class}     & {\em Both} & {\em Seg} & {\em Seg} & {\em Seg} & {\em Seg}  & {\em Class}  \\
    				Number of Cancer Types  & 1 & 8 & 2 & 4 & 1 & 1  \\
    				\bottomrule
    			\end{tabular}
    		}
    	\end{center}
	\end{table*}

	\begin{figure*}[!t]
		\centering
		\includegraphics[width=1.0\textwidth]{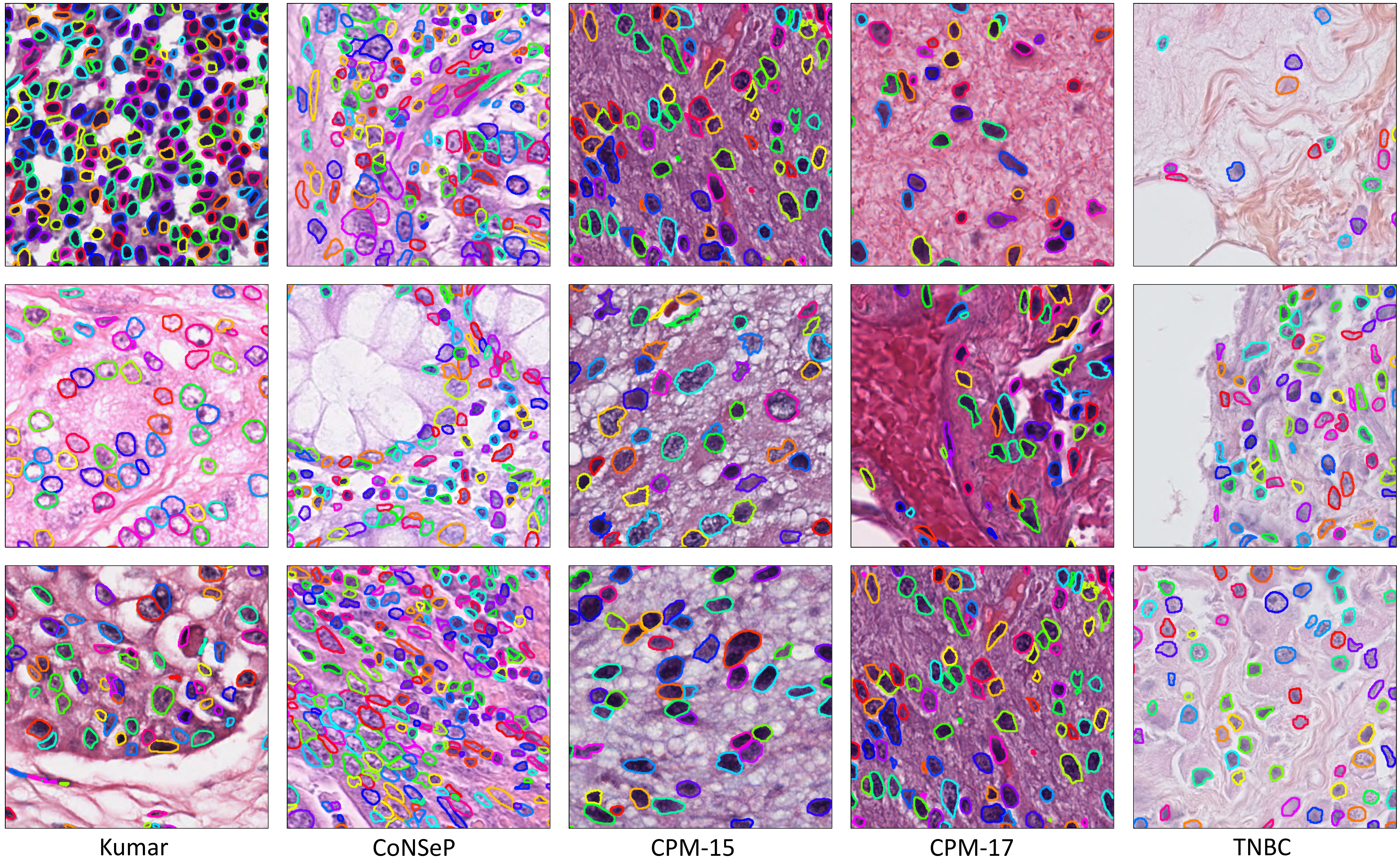} 
		\caption{Sample cropped regions extracted from each of the five nuclear instance segmentation datasets used in our experiments. From left to right: Kumar \cite{kumar}; CoNSeP; CPM-15; CPM-17 \cite{vu2018methods} and TNBC \cite{naylor2018segmentation}. The different colours of nuclear contours highlight individual instances.}
		\label{fig:dataset_samples}
	\end{figure*}
	
	\begin{figure*}[!t]
		\centering
		\includegraphics[width=1.0\textwidth]{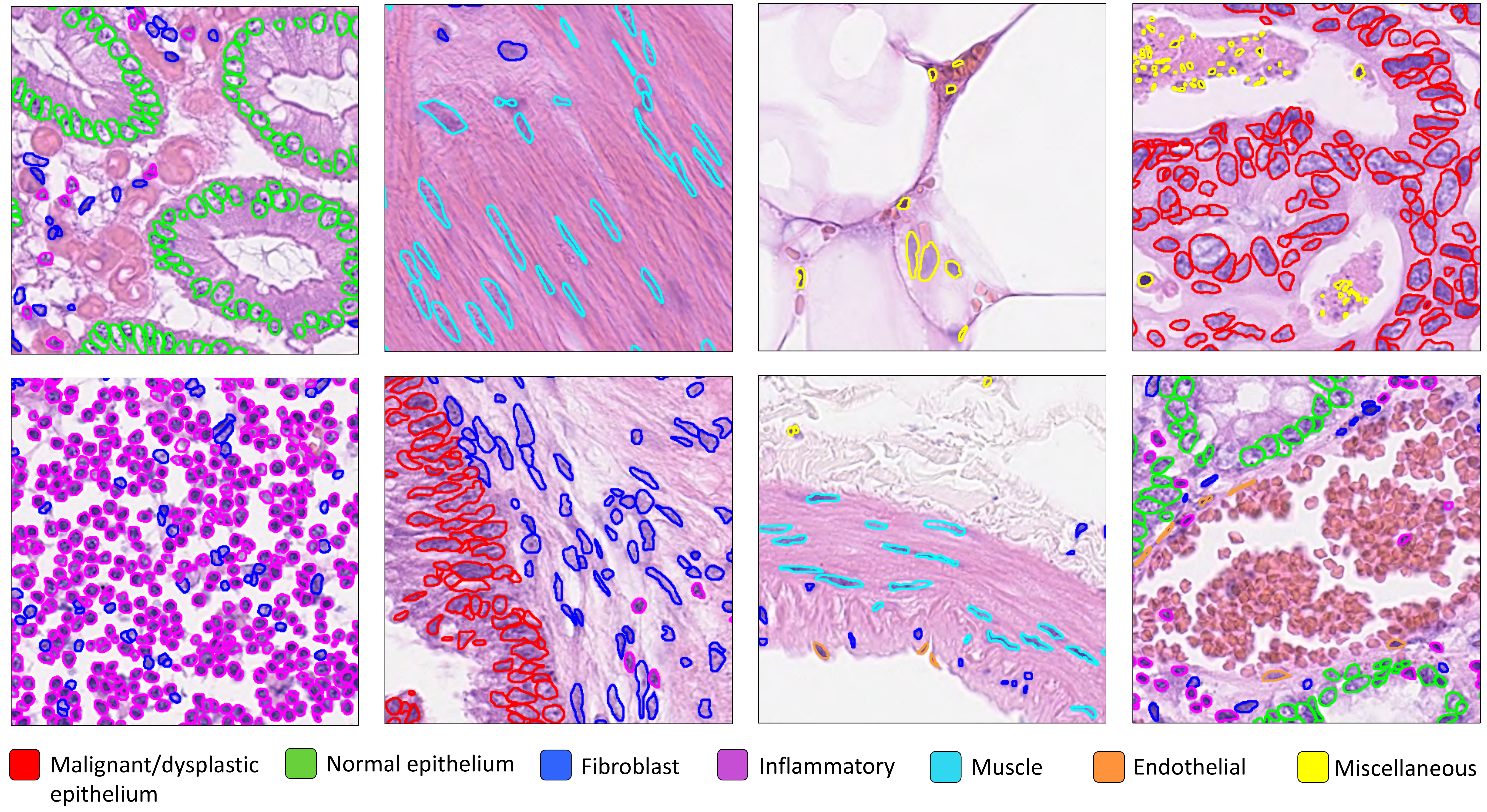} 
		\caption{Sample cropped regions extracted from the CoNSeP datasets, where the colour of each nuclear boundary denotes the category.}
		\label{fig:dataset_samples_class}
	\end{figure*}
	
	As part of this work, we introduce a new dataset that we term as the colorectal nuclear segmentation and phenotypes (CoNSeP) dataset\footnote{This dataset is available at \url{https://warwick.ac.uk/fac/sci/dcs/research/tia/data/}.}, consisting of 41 H\&E stained image tiles, each of size 1,000$\times$1,000 pixels at 40$\times$ objective magnification. Images were extracted from 16 colorectal adenocarcinoma (CRA) WSIs, each belonging to an individual patient, and scanned with an Omnyx VL120 scanner within the department of pathology at University Hospitals Coventry and Warwickshire, UK. We chose to focus on a \textit{single} cancer type, so that we are able to display the true variation of tissue within colorectal adenocarcinoma WSIs, as opposed to other datasets that instead focus on using a small number of visual fields from various cancer types. Within this dataset, stroma, glandular, muscular, collagen, fat and tumour regions can be observed. Beside incorporating different tissue components, the 41 images were also chosen such that different nuclei \textit{types} were present, including: normal epithelial; tumour epithelial; inflammatory; necrotic; muscle and fibroblast. Here, by \textit{type} we are referring to the type of cell from which the nucleus originates from. Within the dataset, there are many significantly overlapping nuclei with indistinct boundaries and there exists various artifacts, such as ink. As a result of the diversity of the dataset, it is likely that a model trained on CoNSeP will perform well for unseen CRA cases. For each image tile, every nucleus was annotated by one of two expert pathologists (A.A, Y-W.T). After full annotation, each annotated sample was reviewed by \textit{both} of the pathologists; therefore refining their own and each others' annotations. By the end of the annotation process, each pathologist had fully checked \textit{every} sample and consensus had been reached. Annotating the data in this way ensured that minimal nuclei were missed in the annotation process. However, we can not avoid inevitable pixel-level differences between the annotation and the true nuclear boundary in challenging cases. In addition to delineating the nuclear boundaries, every nucleus was labelled as either: normal epithelial, malignant/dysplastic epithelial, fibroblast, muscle, inflammatory, endothelial or miscellaneous. Within the miscellaneous category, necrotic, mitotic and cells that couldn't be categorised were grouped. For our experiments, we grouped the normal and malignant/dysplastic epithelial nuclei into a single class and we grouped the fibroblast, muscle and endothelial nuclei into a class named spindle-shaped nuclei. 
	
    Overall, six independent datasets are utilised for this study. A full summary for each of them is provided in Table \ref{table:dataset_summary}. Five of these datasets are used to evaluate the instance segmentation performance which we refer to as: CoNSeP; Kumar \cite{kumar}; CPM-15; CPM-17 \cite{vu2018methods} and TNBC \cite{naylor2018segmentation}. Example images from each of the five datasets can be seen in Fig. \ref{fig:dataset_samples}. Meanwhile, we utilise CoNSeP and a further dataset, named CRCHisto, to quantify the performance of the nuclear classification model. The CRCHisto dataset consists of the same nuclei types that are present in CoNSeP. It is also worth noting that the CRCHisto dataset is not exhaustively annotated for nuclear class labels. 

	\subsection{Implementation and Training Details} \label{section:implementation}
	We implemented our framework with the open source software library TensorFlow version 1.8.0~\cite{abadi2016tensorflow} on a workstation equipped with two NVIDIA GeForce 1080 Ti GPUs. During training, data augmentation including flip, rotation, Gaussian blur and median blur was applied to all methods. All networks received an input patch with a size ranging from 252$\times$252 to 270$\times$270. This size difference is due to the use of valid convolutions in some architectures, such as HoVer-Net and U-Net. Regarding HoVer-Net, we initialised the model with pre-trained weights on the ImageNet dataset \cite{imagenet_cvpr09}, trained only the decoders for the first 50 epochs, and then fine-tuned all layers for another 50 epochs. We train stage one for around 120 minutes and stage two for around 260 minutes. Therefore, the overall training time is around 380 minutes. Stage two takes longer to train because unfreezing the encoder utilises more memory and therefore a smaller batch size needs to be used. Specifically, we used a batch size of 8 and 4 on each GPU for stage one and two respectively. We used Adam optimisation with an initial learning rate of 10$^{-4}$ and then reduced it to a rate of 10$^{-5}$ after 25 epochs. This strategy was repeated for fine-tuning. On the whole, training of the network is stable, where the usage of fully independent decoders helps the network to converge each time. The network was trained with an RGB input, normalised between 0 and 1.

	\newcommand{\STAB}[1]{\begin{tabular}{@{}c@{}}#1\end{tabular}}
	\begin{table*}[t]
		\begin{center}
			\caption{Comparative experiments on the Kumar \cite{kumar}, CoNSeP and CPM-17 \cite{vu2018methods} datasets. WS denotes watershed-based post processing.}
			\label{table:comparative}
			\setlength{\tabcolsep}{3pt} % Default value: 6pt
			\renewcommand{\arraystretch}{1} % Default value: 1
			\resizebox{\textwidth}{!}{
				\begin{tabular}{c|ccccc|ccccc|ccccc}
					\multicolumn{1}{c}{} & \multicolumn{5}{c}{\textbf{Kumar}} & \multicolumn{5}{c}{\textbf{CoNSeP}} & \multicolumn{5}{c}{\textbf{CPM-17}}  \\
					\toprule
 					     {\textbf{Methods}}
						 & \textbf{DICE} & \textbf{AJI} & \textbf{DQ} & \textbf{SQ} & \textbf{PQ} & \textbf{DICE} & \textbf{AJI} & \textbf{DQ} & \textbf{SQ} & \textbf{PQ} & \textbf{DICE} & \textbf{AJI} & \textbf{DQ} & \textbf{SQ} & \textbf{PQ} \\
					\midrule

					%\midrule
					Cell Profiler \cite{carpenter2006cellprofiler}   & 0.623 & 0.366 & 0.423 & 0.704 & 0.300 & 0.434 & 0.202 & 0.249 & 0.705 & 0.179 &    0.570 & 0.338 & 0.368 & 0.702 & 0.261 \\ 
					QuPath \cite{bankhead2017qupath}    & 0.698 & 0.432 & 0.511 & 0.679 & 0.351 & 0.588 & 0.249 & 0.216 & 0.641 & 0.151 & 0.693 & 0.398 & 0.320 & 0.717 & 0.230 \\ 
					FCN8 \cite{long2015fully}        & 0.797 & 0.281 & 0.434 & 0.714 & 0.312 & 0.756 & 0.123 & 0.239 & 0.682 & 0.163 & 0.840 & 0.397 & 0.575 & 0.750 & 0.435 \\ 					
					FCN8 + WS \cite{long2015fully}   & 0.797 & 0.429 & 0.590 & 0.719 & 0.425 & 0.758 & 0.226 & 0.320 & 0.676 & 0.217 & 0.840 & 0.397 & 0.575 & 0.750 & 0.435  \\ 
					SegNet \cite{badrinarayanan2017segnet}      & 0.811 & 0.377 & 0.545 & 0.742 & 0.407 & 0.796 & 0.194 & 0.371 & 0.727 & 0.270 &    0.857 & 0.491 & 0.679 & 0.778 & 0.531 \\ 
					SegNet + WS \cite{badrinarayanan2017segnet}  & 0.811 & 0.508 & 0.677 & 0.744 & 0.506 & 0.793 & 0.330 & 0.464 & 0.721 & 0.335 & 0.856 & 0.594 & 0.779 & 0.784 & 0.614 \\ 
					U-Net \cite{ronneberger2015u}  & 0.758 & 0.556 & 0.691 & 0.690 & 0.478  &  0.724 & 0.482 & 0.488 & 0.671 & 0.328 &   0.813 & 0.643 & 0.778 & 0.734 & 0.578 \\ 
					Mask-RCNN \cite{mrcnn}  & 0.760 & 0.546 & 0.704 & 0.720 & 0.509 & 0.740 & 0.474 & 0.619 & 0.740 & 0.460 &    0.850 & 0.684 & 0.848 & 0.792 & 0.674  \\ 
					%\midrule
					DCAN \cite{chen2016dcan}       & 0.792 & 0.525 & 0.677 & 0.725 & 0.492 & 0.733 & 0.289 & 0.383 & 0.667 & 0.256 &    0.828 & 0.561 & 0.732 & 0.740 & 0.545 \\ 
					Micro-Net \cite{micronet2018Shan}  & 0.797 & 0.560 & 0.692 & 0.747 & 0.519 & 0.794 & 0.527 & 0.600 & 0.745 & 0.449 & 0.857 & 0.668 & 0.836 & 0.788 & 0.661 \\
					DIST \cite{naylor2018segmentation}    & 0.789 & 0.559 & 0.601 & 0.732 & 0.443  & 0.804 & 0.502 & 0.544 & 0.728 & 0.398 &    0.826 & 0.616 & 0.663 & 0.754 & 0.504 \\
					CNN3 \cite{kumar}    & 0.762 & 0.508 & - & - & -  & - & - & - & - & - &    - & - & - & - & - \\
					\midrule
					CIA-Net \cite{zhou2019cia}    & 0.818 & \textbf{0.620} & 0.754 & 0.762 & 0.577  & - & - & - & - & - & - & - & - & - & - \\
					DRAN \cite{vu2018methods}      & - & - & - & - & -  & - & - & - & - & - &  0.862 & 0.683 & 0.811 & 0.804 & 0.657 \\ 
					\midrule
					HoVer-Net       & \textbf{0.826} & 0.618 & \textbf{0.770} & \textbf{0.773} & \textbf{0.597} & \textbf{0.853} & \textbf{0.571} & \textbf{0.702} & \textbf{0.778} & \textbf{0.547} & \textbf{0.869} & \textbf{0.705} & \textbf{0.854} & \textbf{0.814} & \textbf{0.697} \\ 
				
					\bottomrule
				\end{tabular}
			}
		\end{center}
	\end{table*}
	
 	\begin{figure*}[!t]
		\centering
		\includegraphics[width=1.0\textwidth]{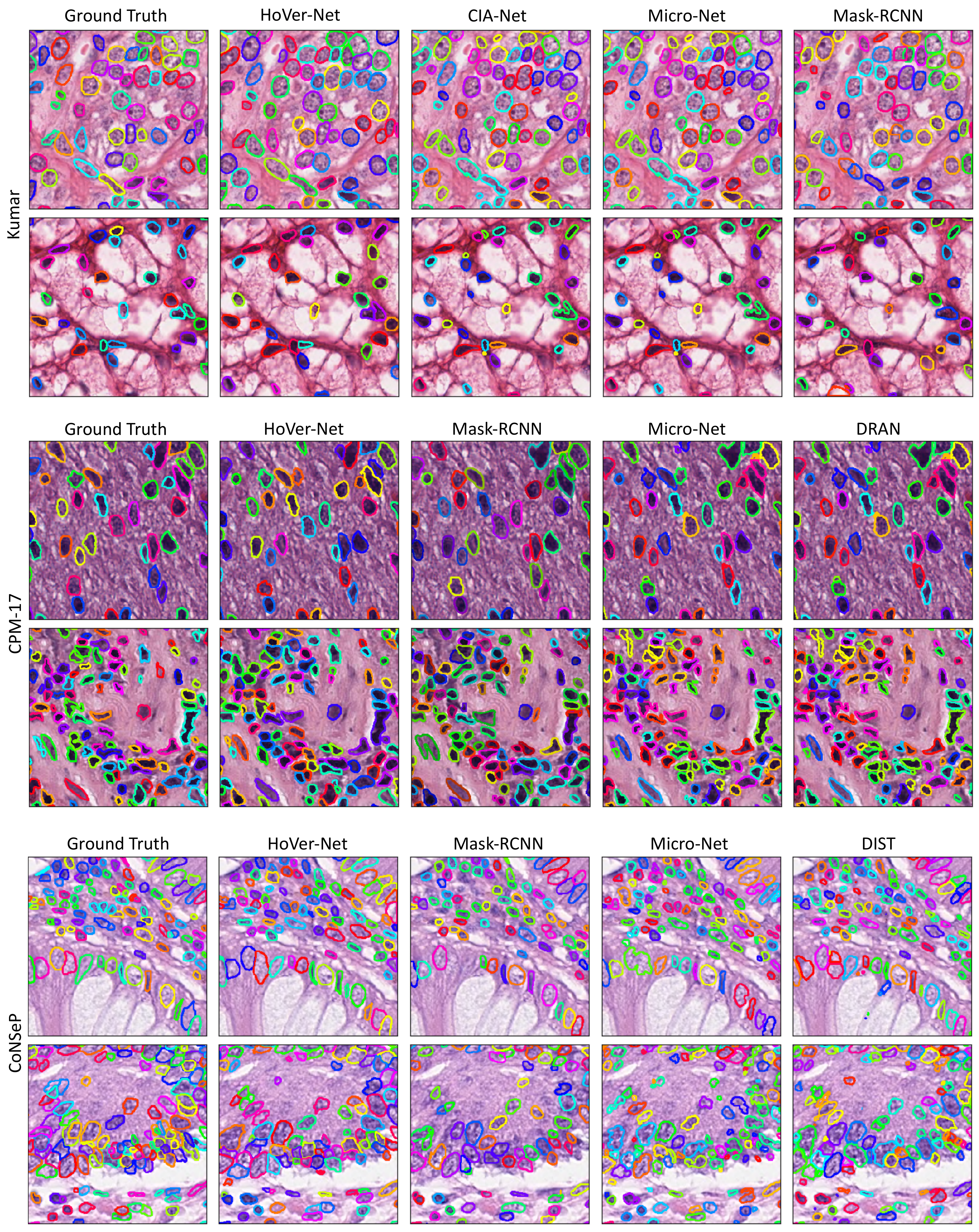} 
		\caption{Example visual results on the CPM-17, Kumar and CoNSeP datasets. For each dataset, we display the 4 models that achieve the highest PQ score from left to right. The different colours of the nuclear boundaries denote separate instances.}
		\label{fig:dataset_samples}
	\end{figure*}
	
	\subsection{Comparative Analysis of Segmentation Methods} \label{section:comparison}

	\textbf{Experimental Setting}:
	We evaluated our approach by employing a full independent comparison across the three largest known exhaustively labelled nuclear segmentation datasets: Kumar; CoNSeP and CPM-17 and utilised the metrics as described in Section \ref{section:metrics_inst}. For this experiment, because we do not have the classification labels for all datasets, we perform instance segmentation without classification. This enables us to fully leverage all data and allows us to rigorously evaluate the segmentation capability of our model. In the same way as \cite{kumar}, we split the Kumar dataset into two different sub-datasets: (i) Kumar-Train, a training set with 16 image tiles (4 breast, 4 liver, 4 kidney and 4 prostate) and (ii) Kumar-Test, a test set with 14 image tiles (2 breast, 2 liver, 2 kidney and 2 prostate, 2 bladder, 2 colon, 2 stomach). Note, we utilise the exact same image split used by other recent approaches \cite{kumar, naylor2018segmentation, zhou2019cia}, but we do not separate the test set into two subsets. We do this to ensure that the test set is large enough, ensuring a reliable evaluation. For CoNSeP, we devise a suitable train and test set that contains 26 and 14 images respectively. The images within the test set were selected to ensure the true diversity of nuclei types within colorectal tissue are represented. For CPM-17, we utilise the same split that had been employed for the challenge, with 32 images in both the training and test datasets.

	We compared our proposed model to recent segmentation approaches used in computer vision \cite{long2015fully, badrinarayanan2017segnet, mrcnn}, medical imaging \cite{ronneberger2015u} and also to methods specifically tuned for the task of nuclear segmentation \cite{chen2016dcan, micronet2018Shan,naylor2018segmentation,zhou2019cia,vu2018methods}. We also compared the performance of our model to two open source software applications: Cell Profiler \cite{carpenter2006cellprofiler} and QuPath \cite{bankhead2017qupath}. Cell Profiler is a software for cell-based analysis, with several suggested pipelines for computational pathology. The pipeline that we adopted applies a threshold to the greyscale image and then uses a series of post processing operations. QuPath is an open source software for digital pathology and whole slide image analysis. To achieve nuclear segmentation, we used the default parameters within the application. FCN, SegNet, U-Net, DCAN, Mask-RCNN and DIST have been implemented by the authors of the paper (S.G, Q.D.V). For Mask-RCNN, we slightly modified the original implementation by using smaller anchor boxes. The default configuration is fine-tuned for natural images and therefore, this modification was necessary to perform a successful nuclear segmentation. DIST was implemented with the assistance of the first author of the corresponding approach in order to ensure reliability during evaluation. This also enabled us to utilise DIST for further comparison in our experiments. For Micro-Net, we used the same implementation that was described by \cite{micronet2018Shan} and was implemented by the first author of the corresponding paper (S.E.A.R). For CNN3 and CIA-Net, we report the results on the Kumar dataset that are given in their respective original papers. The authors of CIA-Net and DRAN provided their segmentation output, which meant that we were able to obtain all metrics on the datasets that the models were applied to. Therefore, we report results of CIA-Net on the Kumar dataset and results of DRAN on the CPM-17 dataset. Note, for all self-implemented approaches we are consistent with our pre-processing strategy. However, DRAN, CNN3 and CIA-Net results are directly taken from their respective papers and therefore we can't guarantee the same pre-processing steps. CNN3 and CIA-Net also use stain normalisation, whereas other methods described in this paper do not. 
	
	\begin{figure*}[!t]
		\centering
        \begin{subfigure}[t]{1.0\linewidth}
            \centering
            \includegraphics[width=0.8\columnwidth]{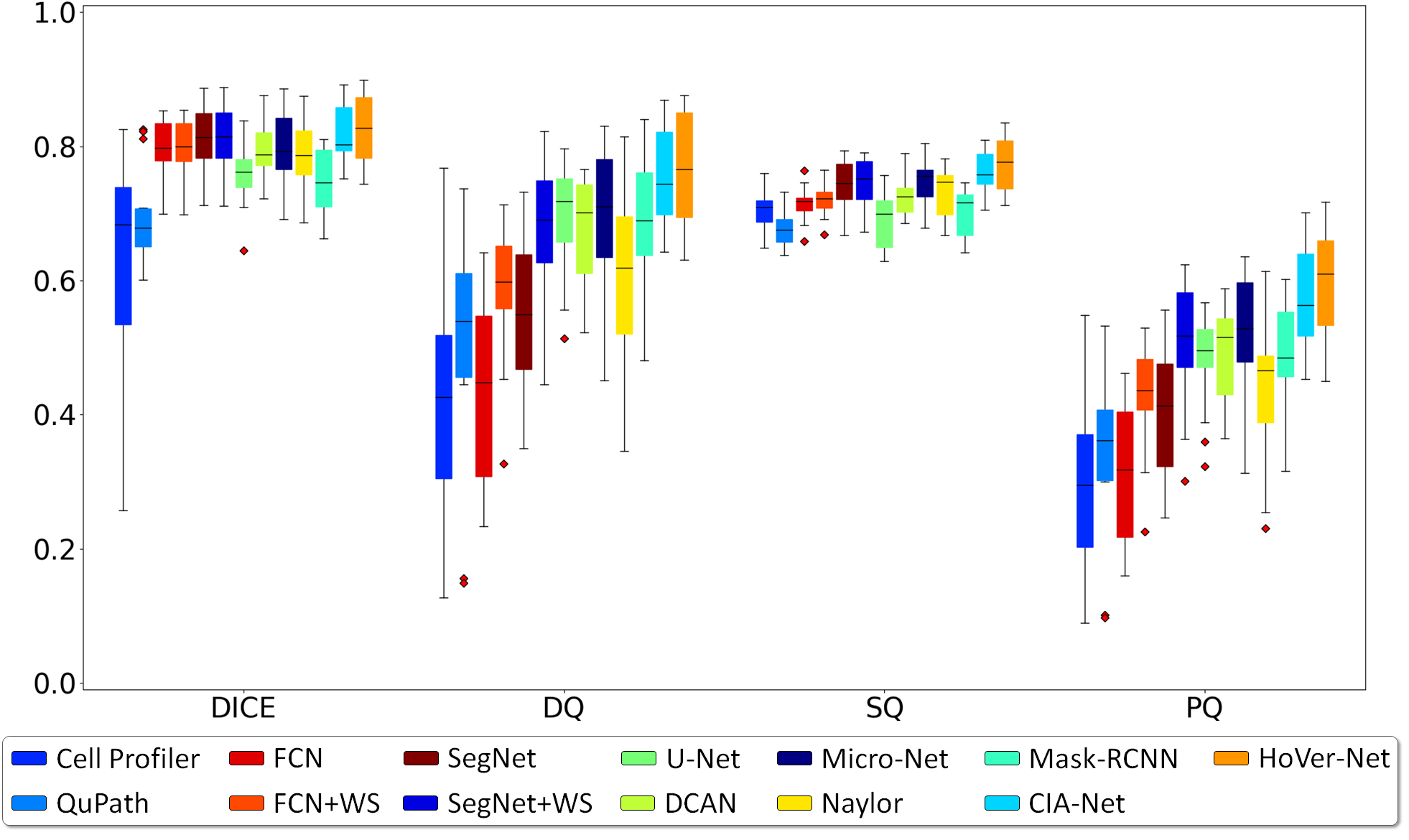} 
            \caption{Kumar}
            \label{fig:boxplot1}
        \end{subfigure}
        \begin{subfigure}[t]{1.0\linewidth}
            \centering
            \includegraphics[width=0.8\columnwidth]{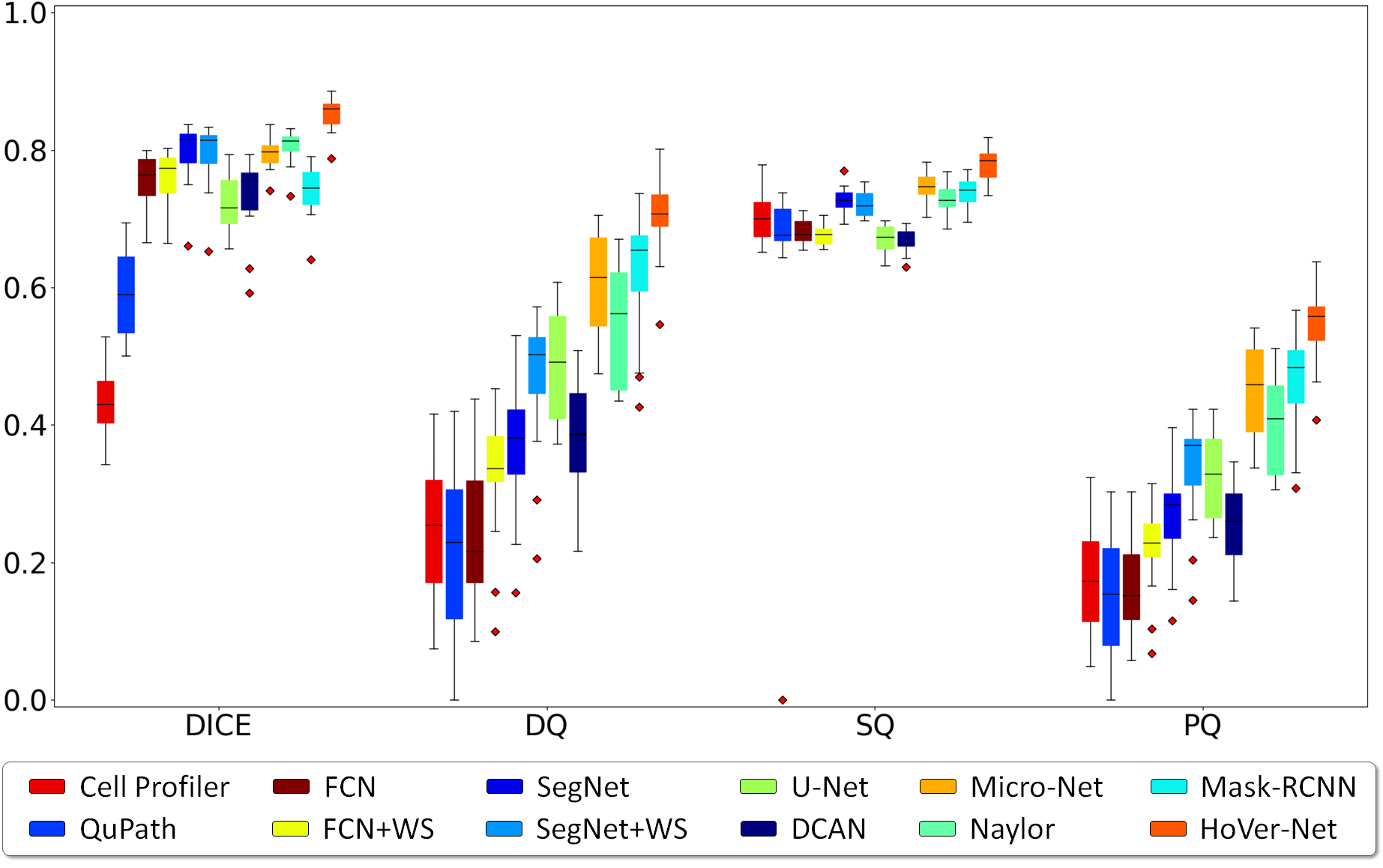} 
            \caption{CoNSeP}
            \label{fig:boxplot2}
        \end{subfigure}
		\caption{Box plots highlighting the performance of competing methods on the Kumar and CoNSeP datasets.}
	\end{figure*}

	\textbf{Comparative Results}: Table \ref{table:comparative} and the box plots in Fig. \ref{fig:boxplot1} and \ref{fig:boxplot2} show detailed results of this experiment. Within the box plots, we choose not to show AJI, due to its limitations as discussed in Section \ref{section:metrics_inst}. A large variation in performance between methods within each dataset is observed. This variation is particularly evident in the Kumar and CoNSeP datasets, where there exists a large number of overlapping nuclei. Both Cell Profiler \cite{carpenter2006cellprofiler} and QuPath \cite{bankhead2017qupath}  achieve sub-optimal performance for all datasets. In particular, both software applications consistently achieve a low DICE score, suggesting that their inability to distinguish nuclear pixels from the background is a major limiting factor. FCN-based approaches improve the capability of models to detect nuclear pixels, yet often fail due to their inability to separate clustered instances. For example, despite a higher DICE score than Cell Profiler and QuPath, networks built only for semantic segmentation like FCN8 and SegNet suffer from low PQ values. Therefore, methods that incorporate strong instance-aware techniques are favourable. Within CPM-17, there are less overlapping nuclei which explains why methods that are not instance-aware are still able to achieve a satisfactory performance. We observe that the weighted cross entropy loss that is used in both U-Net and Micro-Net can help to separate joined nuclei, but its success also depends on the capacity of the network. This is reflected by the increased performance of Micro-Net over U-Net. 

	DCAN is able to better distinguish between separate instances than FCN8, which uses a very similar encoder based on the VGG16 network. Therefore, incorporating additional information at the output of the network can improve the segmentation performance. This is also exemplified by the fairly strong performances of CNN3, DIST, DRAN and CIA-Net. In a different way, Mask-RCNN is able to successfully separate clustered nuclei by utilising a region proposal based approach. However, Mask-RCNN is less effective than other methods at detecting nuclear pixels, which is reflected by a lower DICE score.
	
	Due to the reasoning given in Section \ref{section:metrics}, we place a larger emphasis on PQ to determine the success of different models. In particular, we consistently obtain an improved performance over DIST, which justifies the use of our proposed horizontal and vertical maps as a regression target. We also report a better performance than the winners of the Computational Precision Medicine and MoNuSeg challenges \cite{vu2018methods, zhou2019cia}, that utlised the CPM-17 and Kumar datasets respectively. Therefore, HoVer-Net achieves state-of-the art performance for nuclear instance segmentation compared to all competing methods on multiple datasets that consist of a variety of different tissue types. Our approach also outperforms methods that were fine-tuned for the task of nuclear segmentation. 
	
	\subsection{Generalisation Study} \label{section:generalise}
	
	\textbf{Experimental Setting}: The goal of any automated method is to perform well on unseen data, with high accuracy. Therefore, we conducted a large scale study to assess how all methods generalise to new H\&E stained images. To analyse the generalisation capability, we assessed the ability to segment nuclei from: i) new organs (variation in nuclei shapes) and ii) different centres (variation in staining). 
	 
	The five instance segmentation datasets used within our experiments can be grouped into three groups according to their origin: TCGA (Kumar, CPM-15, CPM-17), TNBC and CoNSeP. We used Kumar as the training and validation set, due to its size and diversity, whilst the combined CPM (CPM-15 and CPM-17), TNBC and CoNSeP datsets were used as three independent test sets. We split the test sets in this way in accordance with their origin. Note, for this experiment we use both the training and test sets of CPM-17 and CoNSeP to form the independent test sets. Kumar was split into three subsets, as explained in Section \ref{section:datasets}, and Kumar-Train was used to train all models, i.e. trained with samples originating from the following organs: breast; prostate; kidney and liver. Despite all samples being extracted from TCGA, CPM samples come from the brain, head \& neck and lungs regions. Therefore, testing with CPM reflects the ability for the model to generalise to new organs, as mentioned above by the first generalisation criterion. TNBC contains samples from an already seen organ (breast), but the data is extracted from an independent source with different specimen preservation and staining practice. Therefore, this reflects the second generalisation criterion. CoNSeP contains samples taken from colorectal tissue, which is not represented in Kumar-Train, and is also extracted from a source independent to TCGA. Therefore, this reflects \textit{both} the first and second generalisation criteria. Also, as mentioned in Section \ref{section:datasets}, CoNSeP contains challenging samples, where there exists various artifacts and there is variation in the quality of slide preparation. Therefore, the performance on this dataset also reflects the ability of a model to generalise to difficult samples.

		\begin{table*}[h]
		\small
		\begin{center}
			\caption{Comparative results, highlighting the generalisation capability of different models. All models are initially trained on Kumar and then the Combined CPM \cite{vu2018methods}, TNBC \cite{naylor2018segmentation} and CoNSeP datasets are processed.}
			\label{table:generalization}
			\setlength{\tabcolsep}{3pt} % Default value: 6pt
			\renewcommand{\arraystretch}{1} % Default value: 1
			\resizebox{\textwidth}{!}{
			\begin{tabular}{c|ccccc|ccccc|ccccc}
			\multicolumn{1}{c}{} & \multicolumn{5}{c}{\textbf{Combined CPM}}  & \multicolumn{5}{c}{\textbf{TNBC}} & \multicolumn{5}{c}{\textbf{All CoNSeP}} \\
    		\midrule
			 {\textbf{Methods}} & \textbf{DICE} & \textbf{AJI} & \textbf{DQ} & \textbf{SQ} & \textbf{PQ}  & \textbf{DICE} & \textbf{AJI} & \textbf{DQ} & \textbf{SQ} & \textbf{PQ}& \textbf{DICE} & \textbf{AJI} & \textbf{DQ} & \textbf{SQ} & \textbf{PQ}\\
				\midrule
				FCN8 + WS \cite{long2015fully}     & 0.762 & 0.531 & 0.669 & 0.722 & 0.487 & 0.726 & 0.506 & 0.662 & 0.723 & 0.480 & 0.609 & 0.247 & 0.345 & 0.688 & 0.240 \\
				SegNet + WS \cite{badrinarayanan2017segnet}     & 0.791 & 0.583 & 0.738 & 0.755 & 0.561 & 0.758 & 0.559 & 0.734 & 0.750 & 0.554 & 0.681 & 0.315 & 0.449 & 0.733 & 0.332 \\
				U-Net \cite{ronneberger2015u}     & 0.720 & 0.541 & 0.652 & 0.672 & 0.446 & 0.681 & 0.514 & 0.635 & 0.676 & 0.442 & 0.585 & 0.363 & 0.442 & 0.670 & 0.297 \\
				Mask-RCNN \cite{mrcnn}        & 0.764 & 0.575 & 0.760 & 0.719 & 0.549 & 0.705 & 0.529 & 0.726 & 0.742 & 0.543 & 0.606 & 0.348 & 0.492 & 0.720 & 0.357 \\
				DCAN \cite{chen2016dcan}     & 0.770 & 0.582 & 0.716 & 0.730 & 0.528 & 0.725 & 0.537 & 0.683 & 0.720 & 0.495 & 0.609 & 0.306 & 0.403 & 0.685 & 0.278 \\
				Micro-Net \cite{micronet2018Shan}   & 0.792 & 0.615 & 0.716 & 0.751 & 0.542 & 0.701 & 0.531 & 0.656 & 0.753 & 0.497 & 0.644 & 0.394 & 0.489 & 0.722 & 0.356 \\
				DIST \cite{naylor2018segmentation}       & 0.775 & 0.563 & 0.593 & 0.720 & 0.432 & 0.719 & 0.523 & 0.549 & 0.714 & 0.404 & 0.621 & 0.369 & 0.379 & 0.701 & 0.268\\
				HoVer-Net      & \textbf{0.801} & \textbf{0.626} & \textbf{0.774} & \textbf{0.778} & \textbf{0.606} & \textbf{0.749} & \textbf{0.590} & \textbf{0.743} & \textbf{0.759} & \textbf{0.578} & \textbf{0.664} & \textbf{0.404} & \textbf{0.529} & \textbf{0.764} & \textbf{0.408} \\
				\bottomrule
			\end{tabular}
			}
		\end{center}
	\end{table*}

	\textbf{Comparative Results}: The results are reported in Table \ref{table:generalization}, where we only display the results of methods that employ an instance-based technique. We observe that our proposed model is able to successfully generalise to unseen data in all three cases. However, some methods prove to perform poorly with unseen data, where in particular, U-Net and DIST perform worse than other competing methods on all three datasets. Both SegNet with watershed and Mask-RCNN achieve a competitive performance across all three generalisation tests. However, similar to the results reported in Table \ref{table:comparative}, Mask-RCNN is not able to distinguish nuclear pixels from the background as well as other competing methods, which has an adverse effect on the overall segmentation performance shown by PQ. On the other hand, SegNet proves to successfully detect nuclear pixels, reporting a greater DICE score than HoVer-Net on both the TNBC and CoNSeP datasets. However, the overall segmentation result for HoVer-Net is superior because it is better able to separate nuclear instances by incorporating the horizontal and vertical maps at the output of the network. 

	\subsection{Comparative Analysis of Classification Methods} \label{section:comparison_class}
	
	\textbf{Experimental Setting}: We converted the top four performing nuclear instance segmentation algorithms, based on their panoptic quality on the CoNSeP dataset, such that they were able to perform simultaneous instance segmentation and classification. As mentioned in Section \ref{section:datasets}, the nuclear categories that we use in our experiments are: miscellaneous, inflammatory, epithelial and spindle-shaped. Specifically, we compared HoVer-Net with Micro-Net, Mask-RCNN and DIST. For Micro-Net, we used an output depth of 5 rather than 2, where each channel gave the probability of a pixel being either background, miscellaneous, inflammatory, epithelial or spindle-shaped. For Mask-RCNN, there is a devoted classification branch that predicts the class of each instance and therefore is well suited to a multi-class setting. DIST performs regression at the output of the network and therefore converting the model such that it is able to classify nuclei into multiple categories is non-trivial. Instead, we add an extra 1$\times$1 convolution at the output of the network that performs nuclear classification. As well as comparing to the aforementioned methods, we compared our approach to a spatially constrained CNN (SC-CNN), that achieves detection and classification. Note, because SC-CNN does not produce a segmentation mask, we do not report the PQ for this method.
	
	\textbf{Comparative Results}: We trained our models on the training set of the CoNSeP dataset and then we evaluated the model on both the test set of CoNSeP and also the entire CRCHisto dataset. Table \ref{table:comprative_class} displays the results of the multi-class models on the CoNSeP and the CRCHisto datasets respectively, where the given metrics are described in Section \ref{section:metrics_class}. For CoNSeP, along with the classification metrics, we provide PQ as an indication of the quality of instance segmentation. However, in CRCHisto, only the nuclear centroids are given and therefore, we exclude PQ from the CRCHisto evaluation because it can't be calculated without the instance segmentation masks. We observe that HoVer-Net achieves a good quality simultaneous instance segmentation and classification, compared to competing methods. It must be noted, that we should expect a lower F$_1$ score for the miscellaneous class because there are significantly less nuclei represented. Also, there is a high diversity of nuclei types that have been grouped within this class, belonging to: mitotic; necrotic and cells that are uncategorisable. Despite this, HoVer-Net is able to achieve a satisfactory performance on this class, where other methods fail. Furthermore, compared to other methods, our approach achieves the best F$_1$ score for epithelial, inflammatory and spindle classes. Therefore, due to HoVer-Net obtaining a strong performance for both nuclear segmentation and classification, we suggest that our model may be used for sophisticated subsequent cell-level downstream analysis in computational pathology.

	\begin{table*}[h]
		\begin{center}
			\caption{Comparative results for nuclear classification on the CoNSeP and CRCHisto datasets. F$_d$ denotes the F$_1$ score for nuclear detection, whereas F$_c^e$, F$_c^i$, F$_c^s$ and F$_c^m$ denote the F$_1$ classification score for the epithelial, inflammatory, spindle-shaped and miscellaneous classes respectively.}
			\label{table:comprative_class}
			\setlength{\tabcolsep}{6pt} % Default value: 6pt
			\renewcommand{\arraystretch}{1} % Default value: 1
			\resizebox{0.9\textwidth}{!}{
			\begin{tabular}{c|cccccc|ccccc}
				\multicolumn{1}{c}{}  & \multicolumn{6}{c}{\textbf{CoNSeP}} & \multicolumn{5}{c}{\textbf{CRCHisto}}  \\
				\midrule

				{\textbf{Methods}} & \textbf{PQ} & \textbf{F$_d$} & \textbf{F$_c^e$} & \textbf{F$_c^i$} & \textbf{F$_c^s$} & \textbf{F$_c^m$} & \textbf{F$_d$} & \textbf{F$_c^e$} & \textbf{F$_c^i$} & \textbf{F$_c^s$} & \textbf{F$_c^m$} \\
				\midrule

                SC-CNN \cite{sirinukunwattana2016locality} & -     &  0.608 & 0.306 & 0.193 & 0.175 & 0.000 & 0.664 & 0.246 & 0.111 & 0.126  & 0.000 \\ 
                DIST \cite{naylor2018segmentation}         & 0.372 &  0.712 & 0.617 & 0.534 & 0.505 & 0.000 & 0.616 & 0.464 & 0.514 & 0.275  & 0.000 \\ 
                Micro-Net \cite{micronet2018Shan}          & 0.430 &  0.743 & 0.615 & 0.592 & 0.532 & 0.117 & 0.638 & 0.422 & 0.518 & 0.249  & 0.059 \\ 
                Mask-RCNN \cite{mrcnn}                     & 0.450 &  0.692 & 0.595 & 0.590 & 0.520 & 0.098 & 0.639 & \textbf{0.503} & 0.537 & 0.294  & 0.077 \\ 
                HoVer-Net          & \textbf{0.516} &  \textbf{0.748} & \textbf{0.635} & \textbf{0.631} & \textbf{0.566} & \textbf{0.426} & \textbf{0.688} & 0.486 & \textbf{0.573} & \textbf{0.302}  & \textbf{0.178} \\ 
				\bottomrule
			\end{tabular}
			}
		\end{center}
	\end{table*}

	\section{Discussion and Conclusions} \label{section:discussion}
	Analysis of nuclei in large-scale histopathology images is an important step towards automated downstream analysis for diagnosis and prognosis of cancer. Nuclear features have been often used to assess the degree of malignancy \cite{gurcan2009histopathological}. However, visual analysis of nuclei is a very time consuming task because there are often tens of thousands of nuclei within a given whole-slide image (WSI). Performing simultaneous nuclear instance segmentation and classification enables subsequent exploration of the role that nuclear features play in predicting clinical outcome. For example, \cite{lu2018nuclear} utilised nuclear features from histology TMA cores to predict survival in early-stage estrogen receptor-positive breast cancer.
	%, but did not look at each nuclear type separately. 
	Restricting the analysis to some specific nuclear types only may be advantageous for accurate analysis in computational pathology.
	%, which is possible using our framework.

	In this paper, we have proposed HoVer-Net for simultaneous segmentation and classification of nuclei within multi-tissue histology images that not only detects nuclei with high accuracy, but also effectively separates clustered nuclei. Our approach has three up-sampling branches: 1) the nuclear pixel branch that separates nuclear pixels from the background; 2) the HoVer branch that regresses the horizontal and vertical distances of nuclear pixels to their centres of mass and 3) the nuclear classification branch that determines the type of each nucleus. We have shown that the proposed approach achieves the state-of-the-art instance segmentation performance compared to a large number of recently published deep learning models across multiple datasets, including tissues that have been prepared and stained under different conditions. This makes the proposed approach likely to translate well to a practical setting due its strong generalisation capacity, which can therefore be effectively used as a prerequisite step before nuclear-based feature extraction. We have shown that utilising the horizontal and vertical distances of nuclear pixels to their centres of mass provides powerful instance-rich information, leading to state-of-the-art performance in histological nuclear segmentation. When the classification labels are available, we show that our model is able to successfully segment and classify nuclei with high accuracy. 
	
	Region proposal (RP) methods, such as Mask-RCNN, show great potential in dealing with overlapping instances because there is no notion of \textit{separating} instances; instead nuclei are segmented independently. However, a major limitation of the RP methods is the difficulty in merging instance predictions between neigbouring tiles during processing. For example, if a sub-segment of a nucleus at the boundary is assigned a label, one must ensure that the remainder of the nucleus in the neighbouring tile is also assigned the same label. To overcome this difficulty, for Mask-RCNN, we utilised an overlapping tile mechanism such that we only considered non-boundary nuclei. 
	
	Regarding the processing time, the average time to process a 1,000$\times$1,000 image tile over 10 runs using Mask-RCNN for segmentation and classification was 106.98 seconds. Meanwhile, HoVer-Net only took an average of 11.04 seconds to complete the same operation; approximately 9.7$\times$ faster. On the other hand, the average processing time for DIST and Micro-Net was 0.600 and 0.832 seconds respectively. Mask-RCNN inherently stores a single instance per channel, which leads to very large arrays in memory when there are many nuclei in a single image patch, which also contributes to the much longer processing time as seen above. Overall, FCN methods seem to better translate to WSI processing compared to Mask-RCNN or RPN methods in general. It must be stressed that the timing is not exact and is dependent on hardware specifications and software implementation. With optimised code and sophisticated hardware, we expect these timings to be considerably different. Additionally, the inference time is also dependent on the size of the output. In particular, with a smaller output size, a smaller stride is also required during processing. For instance, if we used padded convolution in the up-sampling branches of HoVer-Net, then we observe ~5.6$\times$ speed up and the average processing time is 1.97 seconds per 1000$\times$1000 image tile. For fair comparison, all models were processed on a single GPU with 12GB RAM and we fixed the batch size to a size of one. Future work will explore the trade-off between the efficiency of HoVer-Net and its potential to accurately perform instance segmentation and classification. 
	
	A major bottleneck for the development of successful nuclear segmentation algorithms is the limitation of data; particularly with additional associated class labels. In this work, we introduce the colorectal adenocarcinoma nuclear segmentation and phenotypes (CoNSeP) dataset, containing over 24K labelled nuclei from challenging samples to reflect the true difficulty of segmenting nuclei in whole-slide images. Due to the abundance of nuclei with an associated nuclear category, CoNSeP aims to help accelerate the development of further simultaneous nuclear instance segmentation and classification models to further increase the sophistication of cell-level analysis within computational pathology. 
	
	We analysed the common measurements used to assess the true performance of nuclear segmentation models and discussed their limitations. Due to the fact that these measurements did not always reflect the instance segmentation performance, we proposed a set of reliable and informative statistical measures. We encourage researchers to utilise the proposed measures to not only maximise the interpretability of their results, but also to perform a fair comparison with other methods.
	
	Finally, methods have surfaced recently that explore the relationship of various nuclear types within histology images \cite{javed2018cellular, sirinukunwattana2018novel}, yet these methods are limited to spatial analysis because the segmentation masks are not available. Utilising our model for nuclear segmentation and classification enables the exploration of the spatial relationship between various nuclear types combined with nuclear morphological features and therefore may provide additional diagnostic and prognostic value. Currently, our model is trained on a single tissue type, yet due to the strong performance of our instance segmentation model across multiple tissues, we are confident that our model will perform well if we were to incorporate additional tissue types. We observe a low F$_1$ classification score for the miscellaneous category in the classification model because there are significantly less samples within this category and there exists high intra-class variability. Future work will involve obtaining more samples within this category, including necrotic and mitotic nuclei, to improve the class balance of the data.

	\section*{Acknowledgments}
	
	\setcounter{table}{0}
    \renewcommand{\thetable}{A\arabic{table}}
	
	This work was supported by the National Research Foundation of Korea (NRF) grant funded by the Korea government (MSIP) (No. 2016R1C1B2012433) and by the Ministry of Science and ICT (MSIT) (No. 2018K1A3A1A74065728). This work was also supported in part by the UK Medical Research Council (No. MR/P015476/1). NR is part of the PathLAKE digital pathology consortium, which is funded from the Data to Early Diagnosis and Precision Medicine strand of the government’s Industrial Strategy Challenge Fund, managed and delivered by UK Research and Innovation (UKRI). We also acknowledge the financial support from the Engineering and Physical Sciences Research Council and Medical Research Council, provided as part of the Mathematics for Real-World Systems CDT. We thank Peter Naylor for his assistance in the implementation of the DIST network.
	
	\section*{Appendix A. Ablation Studies} 
    
	To gain a full understanding of the contribution of our method, we investigated several of its components. Specifically, we performed the following ablation experiments: (i) contribution of the proposed loss strategy; (ii) Sobel-based post processing technique compared to other strategies and (iii) contribution of the dedicated classification branch. Here, we utilised the Kumar and CoNSeP datasets for (i) and (ii) due to the large number of nuclei present, whereas for (iii) we use CoNSeP and CRCHisto because we do not have the classification labels for Kumar.
 
	\textbf{Loss Terms}: We conducted an experiment to understand the contribution of our proposed loss strategy. First, we used mean squared error (MSE) of the horizontal and vertical distances $L_a$ as the loss function of the HoVer branch and binary cross entropy (BCE) loss $L_c$ as the loss function for the NP branch. We refer to this combination as the \textit{standard} strategy because MSE and BCE are the two most commonly used loss functions for regression and binary classification tasks respectively. Next, we introduced the MSE of the horizontal and vertical gradients $L_b$ to the HoVer branch and the dice loss $L_d$ to the NP branch. The intuition behind our novel $L_b$ is that it enforces the correct structure of the horizontal and vertical map predictions and therefore helps to correctly separate neighbouring instances. The dice loss was introduced because it can help the network to better distinguish between background and nuclear pixels and is particularly useful when there is a class-imbalance. We present the results in Table \ref{table:ablation}, where we observe an increase in all performance measures for our proposed multi-term loss strategy. Therefore, the additional loss terms boost the network's ability to differentiate between nuclear and background pixels (DICE) and separate individual nuclei (DQ and PQ). In particular, there is a significant boost in the SQ for both Kumar and CoNSeP, which suggests that our proposed loss function $L_b$ is necessary to precisely determine where nuclei should be split.
	
	\textbf{Post Processing}: Usually, markers obtained from applying a threshold to an energy landscape (such as the distance map) is enough to provide a competitive input for watershed, as seen by DIST in Table \ref{table:comparative}. Although HoVer-Net is not directly built upon an energy landscape, we devised a Sobel-based method to derive both the energy landscape and the markers. To compare with other methods, we implemented two further techniques for obtaining the energy landscape and the markers. We then exhaustively compared all energy landscape and marker combinations to assess which post processing strategy is the best. We start by linking HoVer to the distance map by calculating the square sum $\chi^2+\varphi^2$, which can be seen as the distance from a pixel to its nearest nuclear centroid. In other words, this is a pseudo distance map. Additionally, $\chi$ and $\varphi$ values can be interpreted as Cartesian coordinates with each nuclear centroid as the origin. By thresholding the values between a certain range, we can obtain the markers. The results of all combinations are shown in Table \ref{table:ablation_post_proc}. Note, our gradient-based post processing technique is specifically designed for the HoVer branch output.
	
	\textbf{Classification Branch}: In order to assess the importance of a devoted branch for concurrent nuclear segmentation and classification, we compared the proposed three branch setup of HoVer-Net to a two branch setup. Here, the two branch setup extends the NP branch to a multi-class setting, by predicting each nuclear type at the output. Then, to obtain the binary mask, the positive channels are combined together after nuclear type prediction. Utilising three branches decouples the tasks of nuclear classification and nuclear detection, where a separate branch is devoted to each task. For this ablation study, we train on the CoNSeP training set and then process both the CoNSeP test set and the entire CRCHisto dataset.
	
	We report results in Table \ref{table:ablation_class}, where we observe that utilising a separate branch devoted to the task of nuclear classification leads to an improved overall performance of simultaneous nuclear instance segmentation and classification in both the CoNSeP and CRCHisto datasets. We can see that if the classification takes place at the output of NP branch, then the network's ability to determine the nuclear type is compromised. This is because the task of nuclear classification is challenging and therefore the network benefits from the introduction of a branch dedicated to the task of classification.

	\begin{table*}[t!]
	\begin{center}
		\caption{Ablation study highlighting the contribution of the proposed loss strategy.}
		\label{table:ablation}
		\setlength{\tabcolsep}{6pt} % Default value: 6pt
		\renewcommand{\arraystretch}{1} % Default value: 1
		\resizebox{0.85\textwidth}{!}{
			\begin{tabular}{c|ccccc|ccccc}
				\multicolumn{1}{c}{} & \multicolumn{5}{c}{\textbf{Kumar}} & \multicolumn{5}{c}{\textbf{CoNSeP}} \\
 				\midrule

    			 \textbf{Strategy} & \textbf{DICE} & \textbf{AJI} & \textbf{DQ}& \textbf{SQ} & \textbf{PQ} & \textbf{DICE} & \textbf{AJI} & \textbf{DQ}& \textbf{SQ} & \textbf{PQ} \\
				\midrule
				
				Standard Loss  & 0.823 & 0.750 & 0.771 & 0.581 & 0.608   & 0.846 & 0.685 & 0.774 & 0.532 & 0.557 \\
				Proposed Loss      & \textbf{0.826} & \textbf{0.770} & \textbf{0.773} & \textbf{0.597} & \textbf{0.618}   & \textbf{0.853} & \textbf{0.702} & \textbf{0.778} & \textbf{0.547} & \textbf{0.571} \\
				\bottomrule
			\end{tabular}
		}
	\end{center}
	\end{table*}
	
	\begin{table*}[t!]
		\begin{center}
			\caption{Ablation study for post processing techniques: Sobel-based versus thresholding to get markers and Sobel-based versus naive conversion to get energy landscape}
			\label{table:ablation_post_proc}
			\setlength{\tabcolsep}{6pt} % Default value: 6pt
			\renewcommand{\arraystretch}{1} % Default value: 1
			\resizebox{0.9\linewidth}{!}{
				\begin{tabular}{c|c|ccccc|ccccc}
				\multicolumn{1}{c}{}  & \multicolumn{1}{c}{}  & \multicolumn{5}{c}{\textbf{Kumar}} & \multicolumn{5}{c}{\textbf{CoNSeP}} \\
					\midrule

					\textbf{Energy} & \textbf{Markers} & \textbf{DICE} & \textbf{AJI} & \textbf{DQ}& \textbf{SQ} & \textbf{PQ} & \textbf{DICE} & \textbf{AJI} & \textbf{DQ} & \textbf{SQ} & \textbf{PQ} \\
					\midrule
					
					$\chi^2+\varphi^2$ & Threshold   & 0.825 & 0.597 & 0.705 & 0.764 & 0.541 & 0.850 & 0.543 & 0.602 & 0.761 & 0.459 \\
					$\chi^2+\varphi^2$ & Sobel       & 0.826 & 0.613 & 0.766 & 0.768 & 0.591 & 0.853 & 0.561 & 0.694 & 0.770 & 0.535 \\
					Sobel     & Threshold   & 0.825 & 0.614 & 0.715 & 0.772 & 0.554 & 0.850 & 0.566 & 0.617 & 0.775 & 0.479 \\
					Sobel     & Sobel       & \textbf{0.826} & \textbf{0.618} & \textbf{0.770} & \textbf{0.773} & \textbf{0.597}  & \textbf{0.853} & \textbf{0.571} & \textbf{0.702} & \textbf{0.778} & \textbf{0.547} \\
					\bottomrule
				\end{tabular}
			}
		\end{center}
	\end{table*}
	
			\begin{table*}[t!]
	\begin{center}
		\caption{Ablation study showing the contribution of the classification branch in HoVer-Net on the CoNSeP dataset. F$_d$ denotes the F$_1$ score for nuclear detection, whereas F$_c^e$, F$_c^i$, F$_c^s$ and F$_c^m$ denote the F$_1$ classification score for the epithelial, inflammatory, spindle-shaped and miscellaneous classes respectively.}
		\label{table:ablation_class}
		\setlength{\tabcolsep}{6pt} % Default value: 6pt
		\renewcommand{\arraystretch}{1} % Default value: 1
		\resizebox{0.9\textwidth}{!}{
			\begin{tabular}{c|cccccc|ccccc}
				\multicolumn{1}{c}{} & \multicolumn{6}{c}{\textbf{CoNSeP}} & \multicolumn{5}{c}{\textbf{CRCHisto}}  \\
			    \midrule
                \textbf{Branches} & \textbf{PQ} & \textbf{F$_d$} & \textbf{F$_c^e$} & \textbf{F$_c^i$} & \textbf{F$_c^s$} & \textbf{F$_c^m$} & \textbf{F$_d$} & \textbf{F$_c^e$} & \textbf{F$_c^i$} & \textbf{F$_c^s$} & \textbf{F$_c^m$} \\
				\midrule

				NP \& HoVer        & 0.499 & 0.736 & \textbf{0.636} & 0.545 & 0.528 & 0.333 & 0.666 & 0.458 & 0.523 & 0.271 & 0.132 \\
				NP \& HoVer \& NC    & \textbf{0.516} &  \textbf{0.748} & 0.635 & \textbf{0.631} & \textbf{0.566} & \textbf{0.426} & \textbf{0.688} & \textbf{0.486} & \textbf{0.573} & \textbf{0.302}  & \textbf{0.178} \\ 
				\bottomrule
			\end{tabular}
		}
	\end{center}
	\end{table*}
		
	\ifCLASSOPTIONcaptionsoff
	\newpage
	\fi

	
	
	%\bibliographystyle{IEEEtran}
	\bibliographystyle{IEEEtran}
	\bibliography{ref.bib}

\begin{thebibliography}{10}
\providecommand{\url}[1]{#1}
\csname url@samestyle\endcsname
\providecommand{\newblock}{\relax}
\providecommand{\bibinfo}[2]{#2}
\providecommand{\BIBentrySTDinterwordspacing}{\spaceskip=0pt\relax}
\providecommand{\BIBentryALTinterwordstretchfactor}{4}
\providecommand{\BIBentryALTinterwordspacing}{\spaceskip=\fontdimen2\font plus
\BIBentryALTinterwordstretchfactor\fontdimen3\font minus
  \fontdimen4\font\relax}
\providecommand{\BIBforeignlanguage}[2]{{%
\expandafter\ifx\csname l@#1\endcsname\relax
\typeout{** WARNING: IEEEtran.bst: No hyphenation pattern has been}%
\typeout{** loaded for the language `#1'. Using the pattern for}%
\typeout{** the default language instead.}%
\else
\language=\csname l@#1\endcsname
\fi
#2}}
\providecommand{\BIBdecl}{\relax}
\BIBdecl

\bibitem{elmore2015diagnostic}
J.~G. Elmore, G.~M. Longton, P.~A. Carney, B.~M. Geller, T.~Onega, A.~N.
  Tosteson, H.~D. Nelson, M.~S. Pepe, K.~H. Allison, S.~J. Schnitt
  \emph{et~al.}, ``Diagnostic concordance among pathologists interpreting
  breast biopsy specimens,'' \emph{Jama}, vol. 313, no.~11, pp. 1122--1132,
  2015.

\bibitem{madabhushi2016digital_pathology}
\BIBentryALTinterwordspacing
A.~Madabhushi and G.~Lee, ``Image analysis and machine learning in digital
  pathology: Challenges and opportunities,'' \emph{Medical Image Analysis},
  vol.~33, pp. 170 -- 175, 2016, 20th anniversary of the Medical Image Analysis
  journal (MedIA). [Online]. Available:
  \url{http://www.sciencedirect.com/science/article/pii/S1361841516301141}
\BIBentrySTDinterwordspacing

\bibitem{alsubaie2018bottom}
N.~Alsubaie, K.~Sirinukunwattana, S.~E.~A. Raza, D.~Snead, and N.~Rajpoot, ``A
  bottom-up approach for tumour differentiation in whole slide images of lung
  adenocarcinoma,'' in \emph{Medical Imaging 2018: Digital Pathology}, vol.
  10581.\hskip 1em plus 0.5em minus 0.4em\relax International Society for
  Optics and Photonics, 2018, p. 105810E.

\bibitem{lu2018nuclear}
C.~Lu, D.~Romo-Bucheli, X.~Wang, A.~Janowczyk, S.~Ganesan, H.~Gilmore, D.~Rimm,
  and A.~Madabhushi, ``Nuclear shape and orientation features from h\&e images
  predict survival in early-stage estrogen receptor-positive breast cancers,''
  \emph{Laboratory Investigation}, vol.~98, no.~11, p. 1438, 2018.

\bibitem{sirinukunwattana2018novel}
K.~Sirinukunwattana, D.~Snead, D.~Epstein, Z.~Aftab, I.~Mujeeb, Y.~W. Tsang,
  I.~Cree, and N.~Rajpoot, ``Novel digital signatures of tissue phenotypes for
  predicting distant metastasis in colorectal cancer,'' \emph{Scientific
  reports}, vol.~8, no.~1, p. 13692, 2018.

\bibitem{javed2018cellular}
S.~Javed, M.~M. Fraz, D.~Epstein, D.~Snead, and N.~M. Rajpoot, ``Cellular
  community detection for tissue phenotyping in histology images,'' in
  \emph{Computational Pathology and Ophthalmic Medical Image Analysis}.\hskip
  1em plus 0.5em minus 0.4em\relax Springer, 2018, pp. 120--129.

\bibitem{corredor2019spatial}
G.~Corredor, X.~Wang, Y.~Zhou, C.~Lu, P.~Fu, K.~Syrigos, D.~L. Rimm, M.~Yang,
  E.~Romero, K.~A. Schalper \emph{et~al.}, ``Spatial architecture and
  arrangement of tumor-infiltrating lymphocytes for predicting likelihood of
  recurrence in early-stage non--small cell lung cancer,'' \emph{Clinical
  Cancer Research}, vol.~25, no.~5, pp. 1526--1534, 2019.

\bibitem{sharma2015multi}
H.~Sharma, N.~Zerbe, D.~Heim, S.~Wienert, H.-M. Behrens, O.~Hellwich, and
  P.~Hufnagl, ``A multi-resolution approach for combining visual information
  using nuclei segmentation and classification in histopathological images.''
  in \emph{VISAPP (3)}, 2015, pp. 37--46.

\bibitem{wang2016automatic}
P.~Wang, X.~Hu, Y.~Li, Q.~Liu, and X.~Zhu, ``Automatic cell nuclei segmentation
  and classification of breast cancer histopathology images,'' \emph{Signal
  Processing}, vol. 122, pp. 1--13, 2016.

\bibitem{yang2006nuclei}
X.~Yang, H.~Li, and X.~Zhou, ``Nuclei segmentation using marker-controlled
  watershed, tracking using mean-shift, and kalman filter in time-lapse
  microscopy,'' \emph{IEEE Transactions on Circuits and Systems I: Regular
  Papers}, vol.~53, no.~11, pp. 2405--2414, 2006.

\bibitem{cheng2009segmentation}
J.~Cheng, J.~C. Rajapakse \emph{et~al.}, ``Segmentation of clustered nuclei
  with shape markers and marking function,'' \emph{IEEE Transactions on
  Biomedical Engineering}, vol.~56, no.~3, pp. 741--748, 2009.

\bibitem{veta2013}
M.~Veta, P.~van Diest, R.~Kornegoor, A.~Huisman, M.~Viergever, and J.~Pluim,
  ``Automatic nuclei segmentation in h\&e stained breast cancer histopathology
  images,'' \emph{PLoS ONE}, vol.~8, no.~7, p. e70221, 2013.

\bibitem{ali2012integrated}
S.~Ali and A.~Madabhushi, ``An integrated region-, boundary-, shape-based
  active contour for multiple object overlap resolution in histological
  imagery,'' \emph{IEEE transactions on medical imaging}, vol.~31, no.~7, pp.
  1448--1460, 2012.

\bibitem{wienert2012detection}
S.~Wienert, D.~Heim, K.~Saeger, A.~Stenzinger, M.~Beil, P.~Hufnagl, M.~Dietel,
  C.~Denkert, and F.~Klauschen, ``Detection and segmentation of cell nuclei in
  virtual microscopy images: a minimum-model approach,'' \emph{Scientific
  reports}, vol.~2, p. 503, 2012.

\bibitem{latorre2013concave}
\BIBentryALTinterwordspacing
A.~LaTorre, L.~Alonso-Nanclares, S.~Muelas, J.~Peña, and J.~DeFelipe,
  ``Segmentation of neuronal nuclei based on clump splitting and a two-step
  binarization of images,'' \emph{Expert Systems with Applications}, vol.~40,
  no.~16, pp. 6521 -- 6530, 2013. [Online]. Available:
  \url{http://www.sciencedirect.com/science/article/pii/S0957417413003904}
\BIBentrySTDinterwordspacing

\bibitem{kwak2015nucleus}
J.~T. Kwak, S.~M. Hewitt, S.~Xu, P.~A. Pinto, and B.~J. Wood, ``Nucleus
  detection using gradient orientation information and linear least squares
  regression,'' in \emph{Medical Imaging 2015: Digital Pathology}, vol.
  9420.\hskip 1em plus 0.5em minus 0.4em\relax International Society for Optics
  and Photonics, 2015, p. 94200N.

\bibitem{liao2016eclipse}
\BIBentryALTinterwordspacing
M.~Liao, Y.~qian Zhao, X.~hua Li, P.~shan Dai, X.~wen Xu, J.~kai Zhang, and
  B.~ji~Zou, ``Automatic segmentation for cell images based on bottleneck
  detection and ellipse fitting,'' \emph{Neurocomputing}, vol. 173, pp. 615 --
  622, 2016. [Online]. Available:
  \url{http://www.sciencedirect.com/science/article/pii/S0925231215011406}
\BIBentrySTDinterwordspacing

\bibitem{litjens2017survey}
G.~Litjens, T.~Kooi, B.~E. Bejnordi, A.~A.~A. Setio, F.~Ciompi, M.~Ghafoorian,
  J.~A. van~der Laak, B.~Van~Ginneken, and C.~I. S{\'a}nchez, ``A survey on
  deep learning in medical image analysis,'' \emph{Medical image analysis},
  vol.~42, pp. 60--88, 2017.

\bibitem{shen2017deep}
D.~Shen, G.~Wu, and H.-I. Suk, ``Deep learning in medical image analysis,''
  \emph{Annual review of biomedical engineering}, vol.~19, pp. 221--248, 2017.

\bibitem{lecun2015deep}
Y.~LeCun, Y.~Bengio, and G.~Hinton, ``Deep learning,'' \emph{nature}, vol. 521,
  no. 7553, p. 436, 2015.

\bibitem{long2015fully}
J.~Long, E.~Shelhamer, and T.~Darrell, ``Fully convolutional networks for
  semantic segmentation,'' in \emph{Proceedings of the IEEE conference on
  computer vision and pattern recognition}, 2015, pp. 3431--3440.

\bibitem{ronneberger2015u}
O.~Ronneberger, P.~Fischer, and T.~Brox, ``U-net: Convolutional networks for
  biomedical image segmentation,'' in \emph{International Conference on Medical
  image computing and computer-assisted intervention}.\hskip 1em plus 0.5em
  minus 0.4em\relax Springer, 2015, pp. 234--241.

\bibitem{micronet2018Shan}
S.~E.~A. {Raza}, L.~{Cheung}, M.~{Shaban}, S.~{Graham}, D.~{Epstein},
  S.~{Pelengaris}, M.~{Khan}, and N.~M. {Rajpoot}, ``{Micro-Net: A unified
  model for segmentation of various objects in microscopy images},''
  \emph{ArXiv e-prints}, p. arXiv:1804.08145, Apr. 2018.

\bibitem{graham2018b}
S.~Graham and N.~M. Rajpoot, ``Sams-net: Stain-aware multi-scale network for
  instance-based nuclei segmentation in histology images,'' in \emph{Biomedical
  Imaging (ISBI 2018), 2018 IEEE 15th International Symposium on}.\hskip 1em
  plus 0.5em minus 0.4em\relax IEEE, 2018, pp. 590--594.

\bibitem{chen2016dcan}
H.~Chen, X.~Qi, L.~Yu, and P.-A. Heng, ``Dcan: deep contour-aware networks for
  accurate gland segmentation,'' in \emph{Proceedings of the IEEE conference on
  Computer Vision and Pattern Recognition}, 2016, pp. 2487--2496.

\bibitem{cui2018deep}
Y.~Cui, G.~Zhang, Z.~Liu, Z.~Xiong, and J.~Hu, ``A deep learning algorithm for
  one-step contour aware nuclei segmentation of histopathological images,''
  \emph{arXiv preprint arXiv:1803.02786}, 2018.

\bibitem{kumar}
N.~Kumar, R.~Verma, S.~Sharma, S.~Bhargava, A.~Vahadane, and A.~Sethi, ``A
  dataset and a technique for generalized nuclear segmentation for
  computational pathology,'' \emph{IEEE Transactions on Medical Imaging},
  vol.~36, no.~7, pp. 1550--1560, July 2017.

\bibitem{khoshdeli2018deep}
M.~Khoshdeli and B.~Parvin, ``Deep leaning models delineates multiple nuclear
  phenotypes in h\&e stained histology sections,'' \emph{arXiv preprint
  arXiv:1802.04427}, 2018.

\bibitem{zhou2019cia}
Y.~Zhou, O.~F. Onder, Q.~Dou, E.~Tsougenis, H.~Chen, and P.-A. Heng, ``Cia-net:
  Robust nuclei instance segmentation with contour-aware information
  aggregation,'' \emph{arXiv preprint arXiv:1903.05358}, 2019.

\bibitem{vu2018methods}
Q.~D. Vu, S.~Graham, M.~N.~N. To, M.~Shaban, T.~Qaiser, N.~A. Koohbanani, S.~A.
  Khurram, T.~Kurc, K.~Farahani, T.~Zhao \emph{et~al.}, ``Methods for
  segmentation and classification of digital microscopy tissue images,''
  \emph{arXiv preprint arXiv:1810.13230}, 2018.

\bibitem{naylor2018segmentation}
P.~Naylor, M.~La{\'e}, F.~Reyal, and T.~Walter, ``Segmentation of nuclei in
  histopathology images by deep regression of the distance map,'' \emph{IEEE
  Transactions on Medical Imaging}, 2018.

\bibitem{mrcnn}
K.~{He}, G.~{Gkioxari}, P.~{Doll{\'a}r}, and R.~{Girshick}, ``{Mask R-CNN},''
  \emph{ArXiv e-prints}, p. arXiv:1703.06870, Mar. 2017.

\bibitem{nguyen2011prostate}
K.~Nguyen, A.~K. Jain, and B.~Sabata, ``Prostate cancer detection: Fusion of
  cytological and textural features,'' \emph{Journal of pathology informatics},
  vol.~2, 2011.

\bibitem{yuan2012quantitative}
Y.~Yuan, H.~Failmezger, O.~M. Rueda, H.~R. Ali, S.~Gr{\"a}f, S.-F. Chin, R.~F.
  Schwarz, C.~Curtis, M.~J. Dunning, H.~Bardwell \emph{et~al.}, ``Quantitative
  image analysis of cellular heterogeneity in breast tumors complements genomic
  profiling,'' \emph{Science translational medicine}, vol.~4, no. 157, pp.
  157ra143--157ra143, 2012.

\bibitem{sirinukunwattana2016locality}
K.~Sirinukunwattana, S.~e~Ahmed~Raza, Y.-W. Tsang, D.~R. Snead, I.~A. Cree, and
  N.~M. Rajpoot, ``Locality sensitive deep learning for detection and
  classification of nuclei in routine colon cancer histology images.''
  \emph{IEEE Trans. Med. Imaging}, vol.~35, no.~5, pp. 1196--1206, 2016.

\bibitem{preact_resnet}
K.~{He}, X.~{Zhang}, S.~{Ren}, and J.~{Sun}, ``{Identity Mappings in Deep
  Residual Networks},'' \emph{ArXiv e-prints}, p. arXiv:1603.05027, Mar. 2016.

\bibitem{imagenet_cvpr09}
J.~Deng, W.~Dong, R.~Socher, L.-J. Li, K.~Li, and L.~Fei-Fei, ``{ImageNet: A
  Large-Scale Hierarchical Image Database},'' in \emph{CVPR09}, 2009.

\bibitem{anurag2018resnet_robustness}
\BIBentryALTinterwordspacing
A.~Arnab, O.~Miksik, and P.~H.~S. Torr, ``On the robustness of semantic
  segmentation models to adversarial attacks,'' \emph{CoRR}, vol.
  abs/1711.09856, 2017. [Online]. Available:
  \url{http://arxiv.org/abs/1711.09856}
\BIBentrySTDinterwordspacing

\bibitem{densenet}
G.~{Huang}, Z.~{Liu}, L.~{van der Maaten}, and K.~Q. {Weinberger}, ``{Densely
  Connected Convolutional Networks},'' \emph{ArXiv e-prints}, p.
  arXiv:1608.06993, Aug. 2016.

\bibitem{kirilov2018panoptic_quality}
\BIBentryALTinterwordspacing
A.~Kirillov, K.~He, R.~B. Girshick, C.~Rother, and P.~Doll{\'{a}}r, ``Panoptic
  segmentation,'' \emph{CoRR}, vol. abs/1801.00868, 2018. [Online]. Available:
  \url{http://arxiv.org/abs/1801.00868}
\BIBentrySTDinterwordspacing

\bibitem{abadi2016tensorflow}
M.~Abadi, P.~Barham, J.~Chen, Z.~Chen, A.~Davis, J.~Dean, M.~Devin,
  S.~Ghemawat, G.~Irving, M.~Isard \emph{et~al.}, ``Tensorflow: A system for
  large-scale machine learning.'' in \emph{OSDI}, vol.~16, 2016, pp. 265--283.

\bibitem{carpenter2006cellprofiler}
A.~E. Carpenter, T.~R. Jones, M.~R. Lamprecht, C.~Clarke, I.~H. Kang,
  O.~Friman, D.~A. Guertin, J.~H. Chang, R.~A. Lindquist, J.~Moffat
  \emph{et~al.}, ``Cellprofiler: image analysis software for identifying and
  quantifying cell phenotypes,'' \emph{Genome biology}, vol.~7, no.~10, p.
  R100, 2006.

\bibitem{bankhead2017qupath}
P.~Bankhead, M.~B. Loughrey, J.~A. Fern{\'a}ndez, Y.~Dombrowski, D.~G. McArt,
  P.~D. Dunne, S.~McQuaid, R.~T. Gray, L.~J. Murray, H.~G. Coleman
  \emph{et~al.}, ``Qupath: Open source software for digital pathology image
  analysis,'' \emph{Scientific reports}, vol.~7, no.~1, p. 16878, 2017.

\bibitem{badrinarayanan2017segnet}
V.~Badrinarayanan, A.~Kendall, and R.~Cipolla, ``Segnet: A deep convolutional
  encoder-decoder architecture for image segmentation,'' \emph{IEEE
  transactions on pattern analysis and machine intelligence}, vol.~39, no.~12,
  pp. 2481--2495, 2017.

\bibitem{gurcan2009histopathological}
M.~N. Gurcan, L.~E. Boucheron, A.~Can, A.~Madabhushi, N.~M. Rajpoot, and
  B.~Yener, ``Histopathological image analysis: A review,'' \emph{IEEE reviews
  in biomedical engineering}, vol.~2, pp. 147--171, 2009.

\end{thebibliography}
	
\end{document}